%% file: main.tex
\documentclass[sigconf, nonacm]{acmart}

\usepackage{xspace}
\usepackage{xcolor}
\usepackage{booktabs} 
\usepackage{tabularx} 
\usepackage{amsmath}  
\usepackage{multirow}
\usepackage{pgfplots}
\usepackage{xcolor}
\usepackage{tabularx}
\usepackage{booktabs}
\usepackage{array}
\usepackage{makecell}
\usepackage{subcaption}

\input{macros}

\author{Rastislav Lenhardt}
\affiliation{%
  \institution{Google}
  \city{Zurich}
  \country{Switzerland}
}
\email{lenhardt@google.com}

\author{Teodora Dobos}
\affiliation{%
  \institution{Technical University of Munich}
  \city{Munich}
  \country{Germany}
}
\affiliation{%
  \institution{Google}
  \city{Zurich}
  \country{Switzerland}
}
\email{teodora.dobos@tum.de}

\author{Thomas Vecchiato}
\affiliation{%
  \institution{University of Copenhagen}
  \city{Copenhagen}
  \country{Denmark}
}
\affiliation{%
  \institution{Google}
  \city{Zurich}
  \country{Switzerland}
}
\email{thomas.vecchiato@di.ku.dk}

\author{Jiří Iša}
\affiliation{%
    \institution{Google}
    \city{Zurich}
    \country{Switzerland}
}
\email{jisa@google.com}

\author{Igor Ginzburg}
\affiliation{%
  \institution{Google}
  \city{San Jose}
  \state{CA}
  \country{USA}
}
\email{igorginzburg@google.com}

\begin{document}
\title{RSLM: Training-Free Vector Quantization for Approximate Nearest Neighbor Search}

\input{sections/abstract}

\maketitle


\input{sections/intro}
\input{sections/related-works}
\input{sections/contributions}
\input{sections/rslm-codec}

\input{sections/evaluation}
\input{sections/results}

\input{sections/results-end-to-end-1908}
\input{sections/discussion}
\input{sections/conclusions}
\input{sections/reporducibility}




\bibliographystyle{ACM-Reference-Format}
\bibliography{biblio}

\end{document}

%% file: macros.tex
\definecolor{teodoracolor}{RGB}{180, 50, 120}

\newcommand{\rslm}{Rslm\xspace}
\newcommand{\kmeans}{k-means\xspace}
\newcommand{\topk}{top-k}

\newcommand{\cmark}{\ding{51}} 
\newcommand{\xmark}{\ding{55}} 

%% file: sections/abstract.tex
\begin{abstract}
By introducing \rslm (Rotated Scaled Lloyd-Max), a family of training-free vector quantization codecs compressing embeddings to 1--4 bits per dimension, we reduce memory cost and memory bandwidth of a typical large-scale Approximate Nearest Neighbor (ANN) search system, while reducing its complexity and keeping or improving recall across multiple benchmark datasets.
State-of-the-art systems filter candidates using coarse partitions, approximately score them to
narrow the set, and then rescore the best with higher precision representations (often $\ge$8 bits per dimension). Our relativized codecs can bring this down to 2--4 bits per dimension.

We use the properties of the ANN system to encode residual vectors instead of full vectors, both for the approximate scoring phase and the rescoring phase. Since Maximum Inner Product Search (MIPS) is very sensitive to vector norms, we correct the $L_2$ norms of quantized vectors. Our major innovation is that we correct the $L_2$ norm of the final reconstructed vector rather than just the residual.
Our rescaling replaces more complicated schemes, such as Anisotropic loss. The residualization scheme gives us a more favorable quality vs size trade-off than generic quantization methods.

Our high-performance implementation leverages a block-wise cascaded Fast Walsh-Hadamard Transform (FWHT) with linear-like complexity, AVX SIMD-optimized codebooks, and a steganographic encoding of scaling factors for perfect cache-line alignment.

\end{abstract}

%% file: sections/intro.tex

\section{Introduction}

\begin{figure}
  \centering
  \includegraphics[width=\linewidth]{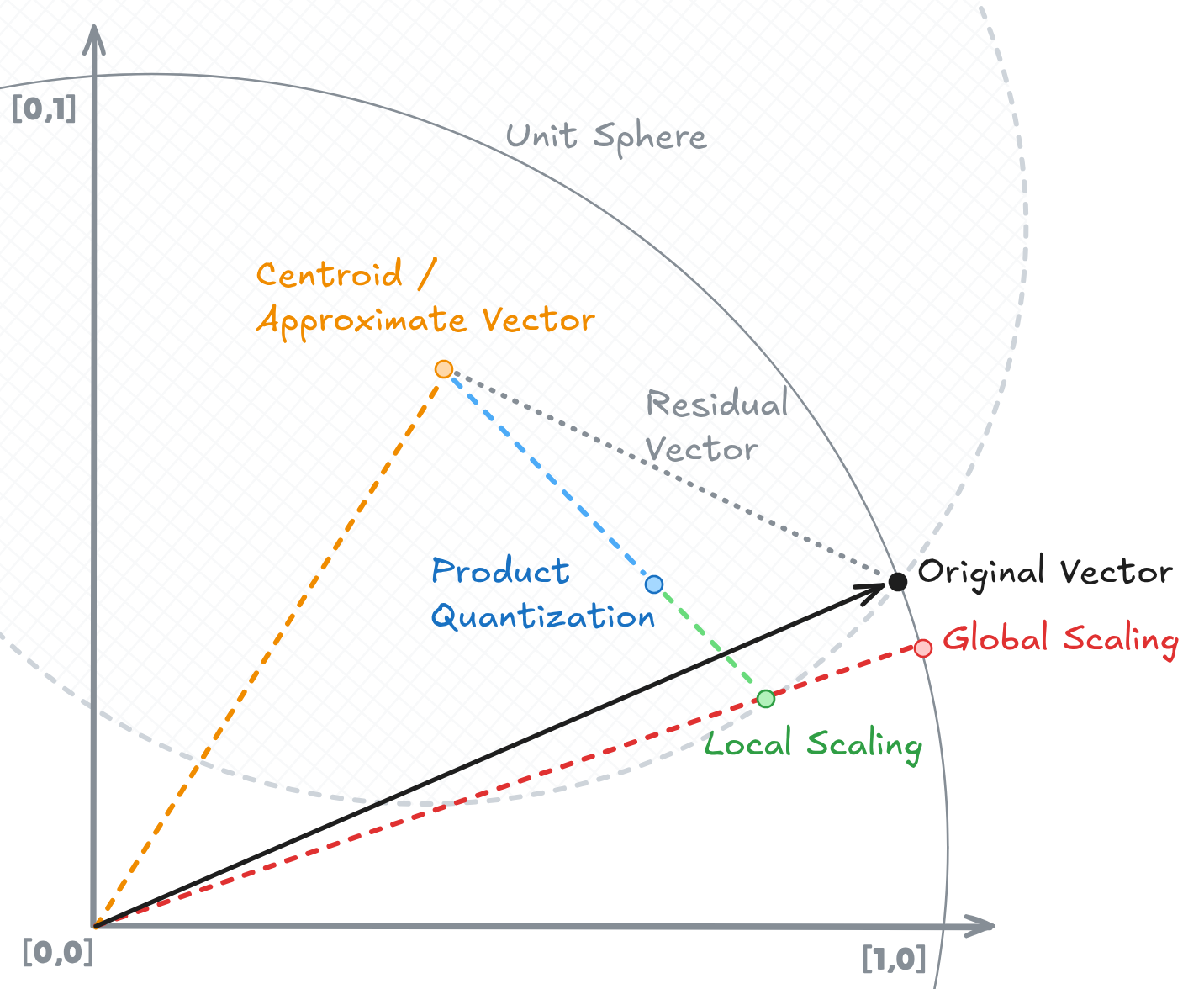}
  \caption{Geometric interpretation of residual quantization with both local and global scaling fixing the norm.}
  \label{fig:resid}
\end{figure}

High-dimensional vector embeddings serve as the foundational representation for capturing unstructured semantic information across various modern domain pipelines, including search~\cite{ma2026vectordb, monir2024vectorsearch, karpukhin2020densepassage}, recommendation~\cite{li2023recommendersystems, raza2025recommendersystems}, and Retrieval-Augmented Generation (RAG)~\cite{sharma2025ragsurvey, gao2024ragsurvey}. Approximate Nearest Neighbor (ANN) search systems are often used to locate relevant vectors within billion-scale databases with high throughput and low latency~\cite{bruch2024vectorretrieval}.
In modern high-throughput database systems, search index design is fundamentally dictated by tight hardware constraints: DRAM-to-CPU memory bandwidth limits and physical DRAM capacity~\cite{jayaram2019diskann}.

High-scale database architectures rely on clustering-based~\cite{jegou2011pq} and tree-based partitioning indexes~\cite{bentley1975tree, dasgupta2013tree}. 
These paradigms divide the vector space into localized Voronoi cells or spatial nodes, restricting online distance evaluations to a pruned subset of candidate vectors. For example, inverted file index (IVF) systems first identify spatially clustered candidate partitions and subsequently perform exhaustive Maximum Inner Product Search (MIPS) candidate re-ranking of candidates within the selected partitions~\cite{vecchiato2024learntrepresentatives, auvolat2015clusteringefficientapproximatemaximum}.

Within this second stage, vector quantization serves as an important technique to compress embeddings into compact representations~\cite{jegou2011pq}. This compression delivers a multi-fold system advantage: it reduces the overall DRAM memory footprint --- enabling billion-scale in-memory indexes --- while simultaneously cutting end-to-end query latency and boosting queries per second throughput by reducing DRAM-to-CPU transfer.

Traditional data-dependent quantization techniques, such as product quantization~\cite{jegou2011pq} (see Section \ref{sec:background_related_work}), rely on offline training to construct dataset-specific codebooks. While data-oblivious quantizers eliminate this offline complexity and overhead, they suffer from recall degradation in aggressive low-bitwidth regimes below $3$ bits per dimension~\cite{gao2025erabitq}. To address this, a recently emerging class of rotation-based data-oblivious quantizers, such as EDEN~\cite{vargaftik2022eden}, TurboQuant~\cite{zandieh2026turboquant} and BlockQuant~\cite{ann2026block}, leverages randomized orthogonal transforms to homogenize coordinate variance prior to quantization. By mapping embedding dimensions to known analytical distributions and using a fixed codebook, these frameworks enable accurate, training-free quantization.

\textit{Our Motivation and Positioning.}
We improve on the existing generic data-oblivious quantizers by respecting ANN-specific constraints and leveraging ANN-specific opportunities. First, dense orthogonal projections (such as the Haar-distributed random orthogonal matrices employed in TurboQuant~\cite{zandieh2026turboquant} and RaBitQ~\cite{gao2024rabitq}) scale quadratically with vector dimensionality $D$, incurring large matrix storage overhead and query transformation latency.
Second, while Fast Walsh-Hadamard Transforms (FWHT) mitigate this computational cost by scaling in $\mathcal{O}(D \log D)$, global full-vector FWHTs are strictly restricted to power-of-two dimensions. When deployed on arbitrary dimensionalities (e.g., $D = 2049$), global FWHT requires zero-padding up to the next power of two, significantly increasing both computational work and memory bandwidth requirements.
Most importantly though, existing general-purpose data-oblivious codecs are traditionally designed as standalone, full-vector encoders optimized for coordinate-wise reconstruction rather than an integrated part of a vector search retrieval system.

\subsection{Contributions}
\label{sec:contributions}

To address these limitations, we introduce \textbf{\rslm (Rotated Scaled Lloyd-Max)}, a training-free vector quantization family of 1- to 4-bit codecs designed for high-performance clustering-based IVF and tree-based ANN rescoring. \rslm is optimized to operate across both the full vector space and the residual space of IVF/tree partitions and/or approximate vectors. 
By quantizing residual vectors, \rslm delivers near-lossless recall at extreme compression rates without any dataset-specific training. Our main contributions follow.


\paragraph{Linear rotation complexity.} 
Instead of dense quadratic projections, or $\mathcal{O}(D \log D)$ FWHT approach by EDEN, \rslm implements a cascaded 2-pass Block-FWHT. By restricting FWHT to fixed-size blocks of size $B$, the formal encoding complexity is $\mathcal{O}(D \log B)$. Since the block size $B$ is a small, hardware-optimized constant (we use $B \le 128$), $\log B$ scales as a constant factor, yielding a linear complexity of $\mathcal{O}(D)$.
    
\paragraph{Final-vector $L_2$ norm correction:} 
MIPS ranking degrades if the norm is not corrected \cite{dai2020norm}. Previous approaches fixed the norm for the residual. \rslm takes this one step further, and explicitly corrects the $L_2$ norm of the final reconstructed vector rather than just the residual, matching or exceeding the recall of complex anisotropic loss formulations without requiring offline codebook training.
    
\paragraph{Zero-byte steganographic metadata:} 
In the 4-bit Rslm4Lite variant, we embed the norm scale factor in the quantized vector. This reduces per-vector metadata overhead to exactly zero and enables perfect cacheline alignment for SIMD register lookups.

\paragraph{ANN simplification with training-free codecs:} 
We evaluate \rslm codecs on full vector quantization and relative quantization in respect to approximate vectors across five datasets, and conduct an end-to-end comparison against Faiss~\cite{douze2025faiss} and ScaNN~\cite{guo2020scann} on  \texttt{glove} and \texttt{openai} datasets, where we employ them to encode approximate vectors. We demonstrate that both in approximate scoring and rescoring phase our training-free codecs match or exceed the quality of the trained codecs.

\paragraph{Memory storage and bandwidth savings:} 
When applied in rescoring phase, our relativized codecs achieve for all five evaluated datasets no recall@20@30 losses with 4-bit quantization (compared to default 8 bits used by ScaNN), and our 2-bit variants achieve 99\%+ recall@20@40.






\subsection{Limitations}

\hspace{\parindent}\emph{Inner Product}. We focus on ANN search to retrieve items with the \textbf{maximum inner product}. Because most modern embedding models output unit-normalized vectors, rankings generated by inner product, cosine similarity, and $L_2$ distance are mathematically identical, and computing dot-products is faster in practice. In our evaluation, four out of five datasets are unit-normalized, while the remaining \texttt{bigann} dataset exhibits minimal norm variance.

\emph{IVF Index}. We focus on tree-based IVF indexes because they are highly optimized for environments with strict physical memory scaling limits and high-throughput requirements. As noted on \textit{ann-benchmarks.com}~\cite{aumuller2020ann}, graph-based indexes like HNSW often achieve lower latency for a single query at a given recall target. However, IVF architectures maintain a crucial advantage in memory footprint and batched processing throughput. For our end-to-end comparison in Subsection \ref{sec:e2e}, we evaluate ScaNN---one of the fastest throughput-optimized algorithms---and Faiss, one of the most widely deployed indexes in industry.

Our codecs can still be very useful also for graph-based HNSW architectures. Non-relative codecs could be used to compress node-vectors to reduce the memory required during graph traversal. And our relative codecs could be applied in a multistage approach similar to Milvus's HNSW-PRQ \cite{wang2021milvus}, where the full precision vectors could be replaced by quantizing residuals against the coarse routing vectors, enabling highly accurate, training-free rescoring.

%% file: sections/related-works.tex
\section{Background and Related Work}
\label{sec:background_related_work}


Scaling ANN search to billions of vectors is constrained by DRAM costs and memory bandwidth~\cite{jayaram2019diskann}. Similarity search engines typically navigate this tradeoff via multi-stage filtering: coarse spatial indices first prune the search space to candidate partitions, followed by quantized in-memory scanning to efficiently rescore vectors.

\subsection{Space Partitioning and Candidate Selection}
Modern similarity search engines decouple candidate pruning from final ranking using space-partitioning index structures~\cite{douze2025faiss}. 
While graph-based architectures like DiskANN~\cite{jayaram2019diskann} achieve exceptional recall-latency trade-offs, their reliance on in-memory graph adjacency lists imposes large physical RAM requirements. 
Similarly, Locality-Sensitive Hashing (LSH) frameworks like Fargo~\cite{zhao2023fargo} are frequently bypassed in production-grade systems due to non-optimal empirical precision and mathematical disalignment with exact MIPS distance properties.
To address these limitations, high-throughput engines adopt a two-stage retrieval pipeline generally built on IVF structures~\cite{jegou2011pq} or tree-based space partitions~\cite{bentley1975tree, dasgupta2013tree}.
For instance, hybrid systems like SPANN~\cite{chen2021spann} store coarse centroids in memory and keep postings on disk, whereas IVF-RaBitQ~\cite{shi2026ivfrabitq} pairs IVF with randomized hypercube projections. 
In the approximate scoring stage, the system first identifies localized Voronoi cells and uses low-bit vector quantization codecs (e.g., 1-bit or 2-bit) to rapidly compute estimated distances and prune candidates under strict memory bandwidth constraints.
In the subsequent rescoring stage, the system fetches full-precision vectors or higher-fidelity quantized representations for a shortlist of \topk ~candidates to re-evaluate exact MIPS distances and finalize the top ranking.

\subsection{Quantization}

Vector quantization (VQ) maps high-dimensional vectors $x \in \mathbb{R}^D$ to a discrete set of codes from a codebook $\mathcal{C} = \{c_1, \dots, c_K\}$ to reduce their memory footprint. The standard objective is to find a quantizer $q(x)$ that minimizes the reconstruction error, typically formulated as the mean squared error (MSE) or squared $L_2$ distortion:
\begin{equation}
\mathcal{L}_{\text{MSE}}(x) = \Vert x - q(x) \Vert^2
\end{equation}

Quantization frameworks can be broadly categorized as either \emph{data-dependent} or \emph{data-oblivious}, depending on how their codebooks and parameters are constructed.

In \emph{data-dependent settings}, codebook learning is performed offline over a training dataset $\mathcal{X}$ by minimizing the empirical reconstruction error using iterative clustering algorithms such as \kmeans. Once learned, online encoding deterministically maps a vector $x$ to its nearest codebook centroid $q(x) = \arg\min_{c \in \mathcal{C}} \|x - c\|^2$. Because the codebook size of unconstrained VQ scales exponentially with the vector dimension $D$ for a fixed quantization error, three main categories of structured codebook relaxations are widely adopted:

\begin{itemize}
\item \emph{Scalar Quantization} (SQ): This approach quantizes each coordinate independently using global or coordinate bounds learned from $\mathcal{X}$ (e.g., the Faiss codecs in Section~\ref{sec:faiss-codecs}). For instance, SQ8 performs scalar quantization to $2^8$ levels, using a static per-dimension scale factor shared across the dataset.

\item \emph{Product Quantization} (PQ)~\cite{jegou2011pq}: This scheme decomposes the \allowbreak\mbox{$D$-dimensional} vector space into $M$ orthogonal subspaces of dimension $D/M$, quantizing each sub-vector independently using a learned sub-codebook $\mathcal{C}^{(m)}$.  Optimized Product Quantization ~\cite{ge2014opq} improves PQ by using a learned orthogonal rotation to balance subspace variances,
while Pyramid-PQ~\cite{wang2026pyramidpd} adaptively merges adjacent subspaces to accelerate online distance computation. 

\item \emph{Multi-Codebook and Residual Quantization} (MCQ/RQ): To achieve higher reconstruction fidelity, multi-codebook methods such as Residual Quantization \cite{chen2010rq}, Product Residual Quantization, and implicit neural codecs like Qinco2 \cite{vallaeys2025qinco2} sequentially encode quantization residuals across cascading stages. However, these approaches require computationally intensive offline training phases and impose substantial decoding overhead during online query execution.
\end{itemize}

In contrast to data-dependent techniques, \textit{data-oblivious methods} eliminate offline training by computing transformation parameters analytically or deriving them per vector at runtime. For instance, \textit{Scaled Scalar Quantization (SSQ)}~\cite{aguerrebere2023similarity} applies per-vector scaling. By storing a scale factor alongside the packed low-bit coordinates,\footnote{In practice, this per-vector scale factor is derived from coordinate min/max bounds \cite{aguerrebere2023similarity}, vector norms \cite{vargaftik2022eden, zandieh2026turboquant}, or scalar multipliers optimized for MSE \cite{vargaftik2022eden, gao2024rabitq}.} SSQ eliminates the need for offline training and guarantees robustness against out-of-distribution data shifts.
We also note that the scale can be hidden inside the quantized vector. For instance, in Section \ref{sec:results} we compare the Rslm codecs against SSQ4Lite, which is a 4-bit scalar quantization codec that hides the 32-bit scale factor within the least significant bits of the quantized vector.\footnote{Specifically, for vectors with at least 64 dimensions, the 32-bit scale factor is embedded directly into the least significant bits of the quantized coordinates.} 

\paragraph{Norm-Aware and MIPS-Oriented Quantization} 
While standard vector quantization minimizes MSE and implicitly weights reconstruction errors uniformly regardless of direction, MIPS depends on vector lengths. Dai et al. \cite{dai2020norm} introduced Norm-Explicit Quantization (NEQ), showing that quantization error in the vector norm has a greater negative impact on inner product ranking than error in vector direction. NEQ explicitly decouples and quantizes the scalar norm separately from the normalized direction vector. Furthermore, ScaNN \cite{guo2020scann} penalizes parallel quantization errors via an anisotropic loss function. These methods demonstrate the necessity of norm preservation for MIPS, but rely on offline codebook training.

\paragraph{Orthogonal Preconditioning and Rotation-Based Codecs.}
While data-oblivious scalar quantizers perform well at high bit levels (e.g., 8-bit), coordinate variance skew and outlier dimensions cause recall degradation in aggressive low-bit regimes. To overcome this without resorting to offline \kmeans training, a distinct class of data-oblivious methods leverages randomized orthogonal preconditioning to homogenize coordinate variance prior to quantization. Within this class, EDEN~\cite{vargaftik2022eden} applies a randomized FWHT over a 1D Gaussian codebook, supporting flexible, non-integer number of bits per coordinate via coordinate sparsification. TurboQuant~\cite{zandieh2026turboquant} constructs a 1D Lloyd-Max codebook calibrated to the exact Beta spherical coordinate marginal under a dense QR rotation, combined with a 1-bit QJL (Quantized Johnson-Lindenstrauss)~\cite{zandieh2025qjl} residual estimator. RaBitQ~\cite{gao2024rabitq} and E-RaBitQ~\cite{gao2025erabitq} project unit-normalized vectors onto rotated hypercube vertices for 1-bit and spherical multi-bit grids, though indexing complexity scales steeply with target bit-rate. BlockQuant~\cite{ann2026block} groups adjacent dimensions to perform continuous spherical \kmeans, relying on dense Haar rotations. 


\paragraph{LLM-Centric Quantization}

Beyond similarity search, vector quantization is used in post-training quantization and activation compression in Large Language Models (LLMs). SmoothQuant ~\cite{xiao2023smoothquant} introduces systematic 8-bit scaling, while QuIP~\cite{chee2023quip} demonstrates that randomized orthogonal preconditioning suppresses channel outliers—a principle generalized by QuaRot~\cite{ashkboos2024quarot} through computational invariance. Advanced low-bit frameworks include CRVQ~\cite{xu2025crvq}, which applies codebook relaxation to critical weight channels, and LiftQuant~\cite{he2026liftquant}, which enables quasi-continuous bit-widths. For memory-efficient attention serving, KIVI~\cite{liu2024kivi} enables 2-bit Key-Value (KV) cache compression, PolarQuant~\cite{wu2025polarquant} applies recursive polar coordinate transformations, and HARP~\cite{zagitov2026harp} leverages learnable butterfly orthogonal transforms. While LLM quantization techniques share preconditioning foundations with search codecs, they optimize for matrix multiplication throughput and perplexity rather than fast MIPS candidate rescoring.

\paragraph{Positioning of Rslm}

As shown in Table~\ref{tab:quantizers_methods}, our proposed \rslm family occupies a unique operating point. We systematically categorize state-of-the-art training-free, rotation-based ANNS quantizers based on three key dimensions: their geometric preconditioning operator, the computational complexity of the projection step, and their per-vector metadata footprint. While classical oblivious quantizers operate exclusively on the full vector space, \rslm is designed to quantize both full vectors and relative residuals relative to coarse tree/IVF partitions. By operating on the residual space, \rslm compresses highly concentrated error vectors, drastically reducing quantization distortion. 

Furthermore, while existing preconditioning methods require either quadratic $\mathcal{O}(D^2)$ dense rotations or super-linear $\mathcal{O}(D \log D)$ projections (as in EDEN~\cite{vargaftik2022eden}), \rslm implements a block-wise cascaded Block-FWHT that enforces a strict linear complexity of $\mathcal{O}(D)$. Finally, while competing schemes rely on explicit per-vector scaling metadata to preserve distance estimation accuracy under inner product search, \rslm scales the reconstructed vector length to preserve the original $L_2$ norm, and through its Rslm4Lite variant, it hides this metadata steganographically, achieving zero storage overhead and enabling highly optimized prefetching.

\begin{table}[H]
\centering
\caption{Taxonomy of state-of-the-art training-free, rotation-based ANNS quantizers, compared with \rslm and Rslm4Lite.}
\label{tab:quantizers_methods}
\footnotesize
\setlength{\tabcolsep}{4pt}
\begin{tabularx}{\columnwidth}{>{\raggedright\arraybackslash}X >{\raggedright\arraybackslash}X c c}
\toprule
\textbf{Method} & \textbf{Rotation} & \shortstack{\textbf{Rotation}\\\textbf{Complexity}} & \textbf{Overhead-Free} \\
\midrule
EDEN \cite{vargaftik2022eden} & Randomized Hadamard & $\mathcal{O}(D \log D)^\dagger$ & \xmark \\
RaBitQ \cite{gao2024rabitq} & Dense Orthogonal Rotation & $\mathcal{O}(D^2)$ & \xmark \\
E-RaBitQ \cite{gao2025erabitq} & Dense Orthogonal Rotation & $\mathcal{O}(D^2)$ & \xmark \\
TurboQuant \cite{zandieh2026turboquant} & Dense Orthogonal Rotation & $\mathcal{O}(D^2)$ & \xmark \\
BlockQuant \cite{ann2026block} & Dense Orthogonal Rotation & $\mathcal{O}(D^2)$ & \xmark \\
\midrule
\rslm (Ours) & Randomized Block-Hadamard & $\mathcal{O}(D)^\dagger$ & \xmark \\
Rslm4Lite (Ours) & Randomized Block-Hadamard & $\mathcal{O}(D)^\dagger$ & \cmark \\
\bottomrule
\multicolumn{4}{l}{\scriptsize $^\dagger$ Hadamard-based transforms are computed via FWHT.}
\end{tabularx}
\end{table}

%% file: sections/contributions.tex


%% file: sections/rslm-codec.tex
\section{Rslm codec family}


In this section, we describe the Rslm codec family, which provides vector quantization across 4, 3, 2, and 1-bit rates.
Each codec in the Rslm family implements three main steps. First, it applies efficient pre-quantization transformations such  that all coordinates have uniform variance. Second, it maps the rotated coordinates onto pre-computed optimal Lloyd-Max codebooks calibrated for the standard normal distribution. Third, it appends a 2-byte per-vector scale to restore vector length and prevent under-scoring in MIPS. 

\subsection{Pre-Quantization Transformations}

Direct quantization of raw embedding vectors can lead to suboptimal codebook capacity utilization due to uneven coordinate variance and outliers. To mitigate this, we follow existing quantization techniques \cite{vargaftik2022eden, zandieh2026turboquant, lee2026orbitquantdataagnosticquantizationimage} and apply orthogonal transformations to the original embeddings. These transformations distribute energy uniformly across all coordinates, shaping the transformed data toward a standard normal distribution while guaranteeing that inner product distances are preserved. 

We implement a fast transformation as a cascaded 2-pass rotation combining three main components: the Fast Walsh-Hadamard Transform (FWHT), static sign-flipping, and block-wise coordinate permutations. These components are detailed below, followed by the complete execution pipeline in Section \ref{sec:cascaded}.

\subsubsection{Fast Walsh Hadamard Transform} \label{sec:hadamard}While applying a full random orthogonal transformation (e.g., via a Haar-distributed random matrix) achieves the desired properties of data uniformization, it is computationally prohibitive. For an embedding vector $x \in \mathbb{R}^D$, general matrix multiplication requires $O(D^2)$ operations. To overcome this computational bottleneck, we employ the Fast Walsh-Hadamard Transform. The unnormalized Walsh-Hadamard matrix $\hat{H}_n \in \{-1, +1\}^{n \times n}$ of dimension $n = 2^k$ is recursively defined as:

\begin{equation}
\begin{aligned}
  \hat{H}_{2^k} &= \begin{bmatrix} 
    \hat{H}_{2^{k-1}} & \hat{H}_{2^{k-1}} \\ 
    \hat{H}_{2^{k-1}} & -\hat{H}_{2^{k-1}} 
  \end{bmatrix}, \\
  \hat{H}_1 &= [1]
\end{aligned}
\end{equation}

The orthonormal transform is given by $H_n = \frac{1}{\sqrt{n}}\hat{H}_n$. The FWHT algorithm exploits this recursive structure to apply the transformation in-place in $O(n \log n)$ time without requiring auxiliary memory. Because the entries of $\hat{H}_n$ are restricted to $\pm 1$, the transformation reduces to a sequence of additions and subtractions, with a single scaling factor of $1/\sqrt{n}$ applied at the end. This avoids expensive multiplications.

To optimize for hardware efficiency and memory bandwidth, we partition the vectors into blocks of constant size $B = 128$ and apply the FWHT to each block independently. This block-wise application bounds the overall computational complexity to $O(D \log B)$—which is linear in the embedding dimension $D$. 
Furthermore, to maximize throughput, we implement the block-wise FWHT using SIMD vector instructions (e.g., AVX2). We optimize memory bandwidth by performing the initial stages of the transform entirely in-register before resolving the remaining stages across registers, maximizing instruction-level parallelism.



To handle arbitrary vector dimensions, we adjust the block-wise application in two ways. First, if the dimension is strictly smaller than 128, we dynamically shrink the block size to the largest power of 2 that fits. Second, if the dimension is not a multiple of the active block size, we use an overlapping approach: we process the vector in standard blocks, and handle the leftover elements by running a final block of the same size that overlaps with the preceding one. Because these block operations are non-commutative, they must be applied in reverse order during vector decompression to achieve exact invertibility.

\subsubsection{Sign Flips and Dimension Permutation} \label{sec:sign-flips-permutation}

The deterministic nature of the Hadamard transform means that it may not sufficiently uniformize certain input distributions (e.g., vectors aligned with the transform axes). To guarantee robust uniformization, we introduce pseudo-randomness using fixed, pre-computed sign-flip and permutation tables. Using static tables ensures that the transformations remain strictly orthogonal while requiring zero metadata overhead.
Specifically, the pipeline utilizes:
\begin{itemize}
    \item \textbf{Sign Flips}: A static table of 256 pseudo-random signs ($\pm 1$) $S_1$ is tiled across the vector coordinates before the transform.
    \item \textbf{Dimension Permutation}: A static 128-element permutation $P$ shuffles coordinates within each block. For blocks smaller than 128, a dynamic bit-reversal permutation is used instead.
\end{itemize}

For vectors exceeding 128 dimensions ($D > 128$), coordinates are interleaved across blocks using a strided transpose to break local block boundaries. For an arbitrary dimension $D$ with active block size $B = 128$, for $M = \lfloor D / B \rfloor$ complete blocks, the prefix of length $M \cdot B$ is permuted and transposed by mapping $x'[k \cdot M + b] = x[b \cdot B + P(k)]$ for $0 \le b < M$ and $0 \le k < B$, while any leftover tail coordinates ($i \ge M \cdot B$) are left unchanged ($x'[i] = x[i]$).
These un-transposed tail coordinates are subsequently rotated and mixed into the global embedding space by the overlapping trailing FWHT block covering indices $[D - B, D)$ in the second pass. This ensures strict orthogonality, exact invertibility, and comprehensive dimension uniformization for arbitrary dimensions without requiring zero-padding or truncation.



\subsubsection{Cascaded Rotation Pipeline} \label{sec:cascaded}
With the individual components defined in Sections \ref{sec:hadamard} and \ref{sec:sign-flips-permutation}, the pre-quantization transformation is executed. 
For the 4-bit codecs (Rslm4 and Rslm4Lite), we apply a single-pass optimization if the vector dimension is small ($D \le 256$). In this case, we only apply the first-pass sign flips and the overlapping FWHT, bypassing the permutation, transpose, and second pass entirely to maximize query throughput.
For larger dimensions ($D > 256$, or always for Rslm1/2/3), the complete transformation is executed in a two-pass cascade. For an input vector $x \in \mathbb{R}^D$, the operations are performed in the following order:
\begin{enumerate}
    \item \textbf{First-Pass Sign Flips}: Multiply coordinates by the static sign-flip sequence $S_1$.
    \item \textbf{First-Pass FWHT}: Apply the block-wise FWHT.
    \item \textbf{Dimension Permutation and Interleaving}: Permute coordinates within each block, and interleave them across blocks using the strided transpose operation.
    \item \textbf{Second-Pass Sign Flips}: Apply the second static sign-flip sequence $S_2$ (which is $S_1$ offset by 127 elements).
    \item \textbf{Second-Pass FWHT}: Apply the final block-wise FWHT.
\end{enumerate}

\subsection{Quantization Using Lloyd-Max Codebook}
\label{sec:quantization}

Following pre-quantization transformations, coordinate values approximate a zero-mean Gaussian distribution ${N}(0, \sigma^2)$~\cite{vargaftik2022eden}. To map these coordinates onto our fixed Lloyd-Max codebooks -- which are calibrated for the standard normal distribution ${N}(0, 1)$ -- we scale the transformed coordinates prior to centroid assignment.

While the theoretical average coordinate variance is given by $\sigma^2 = \frac{1}{D}\|{x}\|^2$, individual rotated vectors might exhibit heavy-tailed coordinate peaks. Scaling strictly by sample variance can cause extreme coordinate outliers to saturate at the codebook boundaries, degrading quantization accuracy. To maintain robust dynamic range control and prevent outlier clipping, we apply Extreme Value Theory (EVT) \cite{coles2001introduction} to compute a vector-adaptive initial scale factor $s_{\text{initial}}$.
Under the assumption that rotated coordinates behave as i.i.d. Gaussians, the expected maximum absolute coordinate value of a $D$-dimensional standard Gaussian vector is a known constant $E_{\max}(D)$. We set $s_{\text{initial}}$ dynamically based on the observed maximum coordinate magnitude of the vector $|z_i|$:
\begin{equation}
s_{\text{initial}} = \frac{\max_i |z_i|}{E_{\max}(D)}
\end{equation}
Because $1/E_{\max}(D)$ is pre-computed, determining $s_{\text{initial}}$ requires only a single pass to locate $\max_i |z_i|$ followed by a scalar multiplication. Normalizing coordinates by $1/s_{\text{initial}}$ dynamically anchors the codebook range to the vector's actual peak coordinate, mitigating tail-clipping errors.

After normalizing the rotated vector by $1/s_{\text{initial}}$, we map its coordinates to the centroids. Depending on the target bit rate $b$, we adapt the codebook dimensionality and centroid count $K$. Specifically, for $b=1$ we use a 4D codebook with $K=16$, for $b=2$ a 2D codebook with $K=16$, and for $b=3$ and $b=4$ we utilize 1D codebooks with $K=8$ and $K=16$ centroids, respectively. When the embedding dimensionality $D$ is not an exact multiple of the sub-vector size ($4$ for $b=1$ or $2$ for $b=2$), the vector is zero-padded to the nearest multiple prior to centroid assignment; during query scoring and norm correction, only the valid $D$ coordinates are accumulated by zeroing out trailing query terms, preserving exact inner-product evaluation without requiring separate 1D fallback codebooks.

As detailed in Section \ref{subsec:performance}, restricting codebooks to at most 16 centroids is critical for hardware SIMD acceleration.
To obtain the fixed Lloyd Max codebooks, we performed \kmeans clustering on a large set of synthetic samples drawn from a standard normal distribution of the corresponding dimensionality using $K$ clusters. The resulting cluster centroids define the optimal quantization levels that minimize the mean squared error under the Gaussian assumption. Importantly, the codebooks that we use are data-independent.

\subsection{Norm Correction} \label{sec:norm-correction}

The Rslm codecs append a 2-byte per-vector scale factor to the quantized vector (except for Rslm4Lite, where the scale factor is  embedded into the first dimensions, see Section~\ref{sec:codecs}) in order to restore vector length and improve MIPS ranking \cite{dai2020norm, guo2020scann}. Concretely, for an  embedding vector $x$ and its first-stage quantization $\hat{x}$, the reconstructed vector representation is $\tilde{x} = s \cdot \hat{x}$. The scale factor $s = \|x\|_2 / \|\hat{x}\|_2$ guarantees that the $L_2$ norm of the reconstructed vector matches that of the original vector.

This exact norm restoration contrasts with standard scaled scalar quantization, which inflates the expected norm, and product quantization, which deflates it (and while doing so they introduce variance in the norm across encoded vectors):
\begin{itemize}
\item SSQ models rounding error as independent uniform noise bounded by the quantization step size $\Delta$. This uniform error over $[-\frac{\Delta}{2}, \frac{\Delta}{2}]$ has a variance of $\sigma^2 = \frac{\Delta^2}{12}$, causing the expected squared magnitude of the quantized vector to inflate according to $\mathbb{E}[\|\hat{x}\|_2^2] \approx \|x\|_2^2 + D \frac{\Delta^2}{12}$ \cite{widrow2008quantization}.
\item PQ centroids represent the conditional means of their corresponding Voronoi cells. Under this centroid property of optimal vector quantization, the expected quantized norm is deflated: $\mathbb{E}[\|q(x)\|_2^2] = \mathbb{E}[\|x\|_2^2] - \text{MSE}$ \cite{gray1998quantization, jegou2011pq}.
\end{itemize}

This theoretical discrepancy is empirically pronounced in high-dimensional spaces. For instance, on the 1536-dimensional \texttt{openai} dataset, where original vectors are unit-normalized, the mean reconstructed norm of SSQ4Lite inflates to $1.82$ ($\sigma = 0.03$), whereas 4-bit per dimension PQ deflates to $0.72$ ($\sigma = 0.01$). Rslm eliminates this distortion entirely by enforcing exact norm alignment.

\subsubsection{Scale Encoding.}

For the scale encoding, Float32 (S1E8M23) offers good range (useful for un-normalized embeddings) and good recall, but 4 bytes is expensive, especially for use cases with lower dimensionality and bits per dimension.  The standard 16-bit floats both have drawbacks:
\begin{itemize}
\item BFloat16 (\texttt{S1E8M7}) offers range similar to Float32, but loses recall.
\item Float16 (\texttt{S1E5M10}) achieves similar recall to Float32, but limits the range.
\end{itemize}
Since scales are non-negative, the sign bit is redundant. A parameter sweep showed \texttt{UE7M9} offering a good middle ground.  We implement \texttt{UE7M9} without subnormals by converting to Float32 with a few simple ALU instructions.

\subsection{Codecs Description} \label{sec:codecs}

All Rslm variants share a unified compression pipeline: 
(1) apply a 2-pass cascaded orthogonal transformation as described in Section \ref{sec:cascaded}, 
(2) estimate the initial scale factor and quantize the normalized vector (Section \ref{sec:quantization}), 
(3) compute the refined scale factor to preserve the $L_2$ norm (Section \ref{sec:norm-correction}), and 
(4) pack the codes and scale factor.

Table~\ref{tab:rslm_codecs} summarizes the technical specifications of each variant.

\begin{table}[ht]
\centering
\caption{Rslm Codec Specifications.}
\label{tab:rslm_codecs}
\begin{tabular}{lcccc}
\hline
\textbf{Codec} & \textbf{Bits/Dim} & \textbf{Type} & \textbf{$K$} & \textbf{Scale} \\ \hline
Rslm1 & 1 & 4D & 16 & Suffix (2B) \\
Rslm2 & 2 & 2D & 16 & Suffix (2B) \\
Rslm3 & 3 & 1D & 8 & Suffix (2B) \\
Rslm4 & 4 & 1D & 16 & Suffix (2B) \\
Rslm4Lite & 4 & 1D & 16/8 & Embedded \\ \hline
\end{tabular}
\end{table}

Rslm4Lite is designed for embedding dimensions $D \ge 64$ to eliminate the 2-byte storage overhead of the scale factor. For the first 16 dimensions, it uses $K=8$ (3-bit) centroids. This frees up the most significant bit in each of the first 16 encoded dimensions. A 16-bit float (\texttt{UE7M9}) scale factor is then steganographically embedded into these 16 unused bits, yielding a packed size of exactly $\lceil D/2 \rceil$ bytes. In practice this means that, e.g., for 256-dimensional embeddings, memory-bandwidth bound systems need only two cachelines instead of three to prefetch the vector. This design was inspired by our optimization run using AlphaEvolve \cite{Novikov2025AlphaEvolve}.

\subsection{Relative Quantization} \label{sec:relative-quantization}

The Rslm codecs can be used either for full vector quantization as described in the previous sections, or for relative quantization (see Figure \ref{fig:resid}). In the latter setting, an Rslm codec is used to quantize the residual vector $r = x - a$, where $x$ is the original vector and $a$ is the vector approximation from the initial quantization stage. \footnote{We discuss how the approximate vector $a$ can be computed in Section \ref{sec:results-relative}.}  
The Rslm family supports two scaling options for relative quantization:

\subsubsection{Local Scaling} 
In the default non-global mode (denoted as RelApprox in Table \ref{tab:recall_eval_relative}), the residual $r$ is quantized on its own to $\tilde{r}$, preserving its own norm via a local scale factor (see Section \ref{sec:norm-correction}). 

To compute the dot-product for ranking, we can reuse the already computed dot-product of the query $q$ with the approximate vector $a$ because
\begin{equation}
    \langle q , (a + \tilde{r}) \rangle = \langle q, a \rangle + \langle q, \tilde{r} \rangle.
\end{equation}
This mode stores one 2-byte (\texttt{UE7M9}) float scale per vector to reconstruct the residual, except for Rslm4Lite, which has zero storage overhead.

\subsubsection{Global Scaling}
To correct for accumulated quantization error across all stages, the global mode (denoted as RelApproxGlobal in Tables \ref{tab:recall_eval_relative} and \ref{tab:recall_eval_relative2040} or Rslm\_Global in Section \ref{sec:e2e}) scales the combined approximation to restore the original vector length. 

We compute a global scale factor $s = \|x\|_2 / \|a + \tilde{r}\|_2$ during encoding and append it as a 2-byte float to the packed data. Thus, two 2-byte floats are stored per vector (see Section \ref{sec:norm-correction}). Since the global scale is close to 1.0, we do not need as much range as the local scale, so we use \texttt{UE4M12}, which devotes more bits to the mantissa, improving recall.  We can achieve zero metadata overhead by applying again the trick from Rslm4Lite also to RelApproxGlobal-Rslm4Lite variant and hiding both scalars in the first 32 dimensions.

During scoring, the final dot product is reconstructed by scaling the combined scores, i.e. equal to $s (\langle q, a \rangle + \langle q, \tilde{r} \rangle)$.


%% file: sections/evaluation.tex
\section{Evaluation}

\subsection{Quality Metrics Used} \label{sec:quality-metrics}

Traditional vector quantization literature evaluates performance via MSE \cite{guo2020scann}. However, in database retrieval, user satisfaction and downstream ranking models depend on ranking accuracy rather than coordinate fidelity. Throughout this paper, we optimize and evaluate against ranking-centric metrics like Recall@X@Y (the fraction of top $X$ ground truth exhaustive search results retrieved at top $Y$ positions). 

We evaluate four metrics: \textbf{Recall@20@30} (our default), \textbf{Recall@20@40} (more lenient, useful for downstream post-processing), \textbf{Recall@20@20} (traditional strict recall), and \textbf{1\%-Tolerant Recall@20} \cite{kuffo2026semanticrecallvectorsearch}. 

Based on findings in the recent industrial benchmarks \cite{kuffo2026semanticrecallvectorsearch}, we use Recall@20@30 as our default. As Table \ref{tab:quality_attribution} illustrates, Recall@20@30 provides a necessary buffer to absorb benign mathematical noise---such as equidistant noise shuffling, tiny quantization artifacts, and borderline order swaps---that strict metrics like traditional Recall@20@20 falsely penalize. Instead, it strictly captures genuine quality failures, such as missing high-relevance matches. While 1\%-Tolerant Recall is conceptually aligned with these goals, Recall@20@30 is significantly easier to measure reliably in distributed database engines, requiring only candidate document IDs to be returned rather than exact uncompressed float scores.

\begin{table}[ht]
    \centering
    \caption{Quality Attribution Across Recall Metrics.}
    \label{tab:quality_attribution}
    \small 
    \renewcommand{\arraystretch}{1.2}
    \begin{tabular}{@{} >{\raggedright\arraybackslash}p{3.2cm} c c c c @{}}
        \toprule
        \textbf{Retrieval Scenario} & \textbf{Valid} & \textbf{R@20} & \textbf{R@20} & \textbf{Tolerant} \\
        & \textbf{Loss?} & \textbf{@20} & \textbf{@30} & \textbf{R@20} \\
        \midrule
        Missing top GT match & Yes & Loss & Loss & Loss \\
        Swapping adjacent items & No & Loss & Ignored & Ignored \\
        Tiny score artifacts & No & Loss & Ignored & Ignored \\
        Dynamic index shifts & No & Loss & Ignored & Ignored \\
        Equidistant noise shuffling & No & Loss & Mitigated & Ignored \\
        \bottomrule
    \end{tabular}
\end{table}

\subsection{Datasets and Brute-Force Evaluation}

We evaluate codecs across five standardized datasets covering diverse dimensionalities, dataset sizes, and coordinate distributions, see Table \ref{tab:datasets}. To decouple quantization loss from index partitioning errors, our quality evaluations focus on brute-force approach. We use 500 queries per dataset and for each query, we isolate the ground-truth top-500 nearest neighbors and evaluate ranking metrics directly on this candidate pool. 

\begin{table}
    \centering
    \caption{Overview of Evaluated Datasets}
    \label{tab:datasets}
    \renewcommand{\arraystretch}{1.2}
    \begin{tabularx}{\linewidth}{@{} l c c X @{}}
        \toprule
        \textbf{Dataset} & \textbf{Dim.} & \textbf{Size} & \textbf{Description \& Characteristics} \\
        \midrule
        \textbf{\texttt{glove}}~\cite{pennington2014glove}     & 100  & 1 M   & Word embeddings \\
        \textbf{\texttt{bigann}}~\cite{jegou2011bigann}    & 128  & 100 M & SIFT image descriptors; strictly non-negative coordinates with strong inter-dimension correlation; not unit normalized \\
        \textbf{\texttt{openai}}~\cite{Cohan2018}    & 1536 & 2 M   & OpenAI text embeddings \\
        \textbf{\texttt{wiki\_full}} & 3072 & 2 M   & Gemini embeddings of Wikipedia \\
        \textbf{\texttt{wiki\_pca}}  & 384  & 2 M   & PCA + random rotation applied to \texttt{wiki\_full} \\
        \bottomrule
    \end{tabularx}
\end{table}

As we are interested in codecs with limited quantization error, 
it is improbable for a candidate outside the top-500 ground truth to reach the top 20 or 30 with this limited error.
We verified this also in practice for \texttt{openai}, \texttt{bigann} and \texttt{glove} datasets by comparing with quantizing all documents for given queries (see our reproduction Colab notebook for \texttt{glove}). The only statistically significant impact was recall difference on non-relativized Rslm1 for all and Rslm2 only for \texttt{bigann}, i.e., when the recall was already low. Using the best-scoring subset for the evaluation allowed us to both iterate quickly and be closer to the real world settings, where modern ANN systems usually score only a small fraction of all the results.

\subsection{Evaluated codecs}
We compare our \rslm codecs with SQ8 (default rescoring codec used in ScaNN), SSQ4Lite (simple 4-bit scalar quantization variant which doesn't need a codebook), and 4-bit Turboquant representing as a reference point EDEN, TurboQuant and BlockQuant (several existing algorithms on which ideas we build). We specifically picked TurboQuant as we had access to optimized C++ version of its code.

%% file: sections/results.tex
\section{Results} \label{sec:results}

\subsection{Quality Results for Full Vector Quantization}

\begin{table*}[ht] 
    \centering
    \caption{Recall@20@30 Evaluation Results for Non-Relative Codecs.}
    \label{tab:recall_eval_non_relative}
    \renewcommand{\arraystretch}{1.2}
    \begin{tabular}{@{} l c c c c c c @{}}
        \toprule
        \textbf{Codec} & \textbf{Bits/Dim} & \textbf{\texttt{wiki\_full}} & \textbf{\texttt{wiki\_pca}} & \textbf{\texttt{bigann}} & \textbf{\texttt{glove}} & \textbf{\texttt{openai}} \\
        & & ($D=3072$) & ($D=384$) & ($D=128$) & ($D=100$) & ($D=1536$) \\
        \midrule
        SQ8        & 8   & \textbf{100.0\%} \textcolor{gray}{$\scriptstyle _{-0.0\%}^{+0.0\%}$} & \textbf{100.0\%} \textcolor{gray}{$\scriptstyle _{-0.0\%}^{+0.0\%}$} & 98.8\% \textcolor{gray}{$\scriptstyle _{-0.5\%}^{+0.4\%}$} & \textbf{100.0\%} \textcolor{gray}{$\scriptstyle _{-0.0\%}^{+0.0\%}$} & \textbf{99.6\%} \textcolor{gray}{$\scriptstyle _{-0.2\%}^{+0.1\%}$} \\
        TurboQuant & 4+s & \textbf{100.0\%} \textcolor{gray}{$\scriptstyle _{-0.0\%}^{+0.0\%}$} & \textbf{99.3\%} \textcolor{gray}{$\scriptstyle _{-0.2\%}^{+0.2\%}$} & 92.2\% \textcolor{gray}{$\scriptstyle _{-0.7\%}^{+0.6\%}$} & 96.9\% \textcolor{gray}{$\scriptstyle _{-0.4\%}^{+0.4\%}$} & \textbf{99.3\%} \textcolor{gray}{$\scriptstyle _{-0.2\%}^{+0.1\%}$} \\
        Rslm4      & 4+s & \textbf{99.9\%} \textcolor{gray}{$\scriptstyle _{-0.1\%}^{+0.1\%}$} & \textbf{99.2\%} \textcolor{gray}{$\scriptstyle _{-0.2\%}^{+0.2\%}$} & 92.3\% \textcolor{gray}{$\scriptstyle _{-0.7\%}^{+0.6\%}$} & 97.3\% \textcolor{gray}{$\scriptstyle _{-0.4\%}^{+0.4\%}$} & 98.9\% \textcolor{gray}{$\scriptstyle _{-0.2\%}^{+0.2\%}$} \\
        Rslm4Lite  & 4   & \textbf{100.0\%} \textcolor{gray}{$\scriptstyle _{-0.0\%}^{+0.0\%}$} & \textbf{99.2\%} \textcolor{gray}{$\scriptstyle _{-0.2\%}^{+0.2\%}$} & 89.5\% \textcolor{gray}{$\scriptstyle _{-0.9\%}^{+0.8\%}$} & 95.5\% \textcolor{gray}{$\scriptstyle _{-0.5\%}^{+0.5\%}$} & \textbf{99.0\%} \textcolor{gray}{$\scriptstyle _{-0.2\%}^{+0.2\%}$} \\
        Slm4       & 4+s & 79.1\% \textcolor{gray}{$\scriptstyle _{-0.9\%}^{+1.0\%}$} & 98.6\% \textcolor{gray}{$\scriptstyle _{-0.2\%}^{+0.2\%}$} & 60.9\% \textcolor{gray}{$\scriptstyle _{-1.7\%}^{+1.6\%}$} & 93.6\% \textcolor{gray}{$\scriptstyle _{-0.6\%}^{+0.6\%}$} & 57.5\% \textcolor{gray}{$\scriptstyle _{-1.4\%}^{+1.3\%}$} \\
        SSQ4Lite   & 4   & 89.8\% \textcolor{gray}{$\scriptstyle _{-0.8\%}^{+0.7\%}$} & 98.1\% \textcolor{gray}{$\scriptstyle _{-0.3\%}^{+0.3\%}$} & 60.2\% \textcolor{gray}{$\scriptstyle _{-1.6\%}^{+1.4\%}$} & 93.3\% \textcolor{gray}{$\scriptstyle _{-0.6\%}^{+0.5\%}$} & 60.6\% \textcolor{gray}{$\scriptstyle _{-1.2\%}^{+1.3\%}$} \\
        Rslm3      & 3+s & \textbf{99.0\%} \textcolor{gray}{$\scriptstyle _{-0.2\%}^{+0.2\%}$} & 94.6\% \textcolor{gray}{$\scriptstyle _{-0.5\%}^{+0.4\%}$} & 75.3\% \textcolor{gray}{$\scriptstyle _{-1.1\%}^{+1.0\%}$} & 87.2\% \textcolor{gray}{$\scriptstyle _{-0.9\%}^{+0.8\%}$} & 94.7\% \textcolor{gray}{$\scriptstyle _{-0.5\%}^{+0.5\%}$} \\
        Rslm2      & 2+s & 94.2\% \textcolor{gray}{$\scriptstyle _{-0.5\%}^{+0.5\%}$} & 82.2\% \textcolor{gray}{$\scriptstyle _{-0.9\%}^{+0.8\%}$} & 49.7\% \textcolor{gray}{$\scriptstyle _{-1.2\%}^{+1.2\%}$} & 67.9\% \textcolor{gray}{$\scriptstyle _{-1.4\%}^{+1.4\%}$} & 82.2\% \textcolor{gray}{$\scriptstyle _{-1.0\%}^{+0.9\%}$} \\
        Rslm1      & 1+s & 79.2\% \textcolor{gray}{$\scriptstyle _{-0.9\%}^{+0.9\%}$} & 56.2\% \textcolor{gray}{$\scriptstyle _{-1.2\%}^{+1.1\%}$} & 25.4\% \textcolor{gray}{$\scriptstyle _{-0.9\%}^{+0.9\%}$} & 39.2\% \textcolor{gray}{$\scriptstyle _{-1.6\%}^{+1.7\%}$} & 61.0\% \textcolor{gray}{$\scriptstyle _{-1.4\%}^{+1.2\%}$} \\
        \bottomrule
    \end{tabular}

    \vspace{0.5em}
    \small{\textit{Note}: `+s' indicates that an extra scaling factor is stored per vector.}
\end{table*}

We present the key observations based on Table \ref{tab:recall_eval_non_relative}, which shows Recall@20@30 on full vector (non-relative) quantization with 95\% confidence intervals. 

\begin{enumerate}
    \item \textbf{Impact of Rotation}: 
    The unrotated codecs Slm4\footnote{Slm4 (Scaled Lloyd-Max) is a codec similar to Rslm4, but without the rotation.} and \allowbreak\mbox{SSQ4Lite} suffer severe recall drops on \texttt{wiki\_full} (79.1\% and 89.8\%), \texttt{bigann} (60.9\% and 60.2\%), and \texttt{openai} (57.5\% and 60.6\%) due to non-zero coordinate means and variance skew.
    In contrast, Rslm4 and TurboQuant achieve >99\% recall on \texttt{wiki\_full} and \texttt{openai} and >92\% recall on \texttt{bigann} by homogenizing variance prior to quantization. \texttt{wiki\_pca} had full random rotation applied as the last step of PCA, so Slm4 and SSQ4Lite perform there well as expected.
    \item \textbf{Dimensionality Scaling}: Higher embedding dimensionality naturally protects against quantization noise.  On \allowbreak\mbox{\texttt{wiki\_full}} (D=3072), even 3-bit and 2-bit codecs achieve 99\% and 94\% recall, respectively. These results are supported by the theoretical properties of the behavior of random projections and quantization noise \cite{widrow1956study, bingham2001random}. The true inner product signal power grows quadratically, while the variance of uncorrelated, zero-mean quantization noise grows linearly. Consequently, the Signal-to-Noise Ratio of the estimated inner product scales as $D \cdot 2^{2b}$ \cite{guo2020scann}. Therefore, on average, quadrupling the dimensionality allows a reduction of one bit per dimension without degrading the statistical reliability of the inner product.
    \item \textbf{Lite vs. Standard 4-bit}: Embedding the scalar avoids per-vector storage overhead for cacheline alignment (see Subsection \ref{subsec:performance}). Although this trade-off is more visible on lower-dimensional datasets where the 16 hidden dimensions represent a larger fraction of the vector, the impact remains negligible in high dimensions, making Rslm4Lite optimal for high-dimensional workloads.
\end{enumerate}

\subsection{Quality Results for Residual (Relative) Vector Quantization} \label{sec:results-relative}

\begin{table*}[ht] 
    \centering
    \caption{Recall@20@30 Evaluation Results for Relative Codecs.}
    \label{tab:recall_eval_relative}
    \renewcommand{\arraystretch}{1.2}
    \begin{tabular}{@{} l c c c c c c @{}}
        \toprule
        \textbf{Codec} & \textbf{Bits/Dim} & \textbf{\texttt{wiki\_full}} & \textbf{\texttt{wiki\_pca}} & \textbf{\texttt{bigann}} & \textbf{\texttt{glove}} & \textbf{\texttt{openai}} \\
        & & ($D=3072$) & ($D=384$) & ($D=128$) & ($D=100$) & ($D=1536$) \\
        \midrule
        RelApprox-SQ8             & 8   & \textbf{100.0\%} \textcolor{gray}{$\scriptstyle _{-0.0\%}^{+0.0\%}$} & \textbf{100.0\%} \textcolor{gray}{$\scriptstyle _{-0.0\%}^{+0.0\%}$} & \textbf{99.9\%} \textcolor{gray}{$\scriptstyle _{-0.1\%}^{+0.1\%}$} & \textbf{100.0\%} \textcolor{gray}{$\scriptstyle _{-0.0\%}^{+0.0\%}$} & \textbf{100.0\%} \textcolor{gray}{$\scriptstyle _{-0.0\%}^{+0.0\%}$} \\
        RelApproxGlobal-Rslm4     & 4+s & \textbf{100.0\%} \textcolor{gray}{$\scriptstyle _{-0.0\%}^{+0.0\%}$} & \textbf{100.0\%} \textcolor{gray}{$\scriptstyle _{-0.0\%}^{+0.0\%}$} & \textbf{100.0\%} \textcolor{gray}{$\scriptstyle _{-0.0\%}^{+0.0\%}$} & \textbf{100.0\%} \textcolor{gray}{$\scriptstyle _{-0.0\%}^{+0.0\%}$} & \textbf{100.0\%} \textcolor{gray}{$\scriptstyle _{-0.0\%}^{+0.0\%}$} \\
        RelApproxGlobal-Rslm4Lite & 4   & \textbf{100.0\%} \textcolor{gray}{$\scriptstyle _{-0.0\%}^{+0.0\%}$} & \textbf{100.0\%} \textcolor{gray}{$\scriptstyle _{-0.0\%}^{+0.0\%}$} & \textbf{99.9\%} \textcolor{gray}{$\scriptstyle _{-0.1\%}^{+0.0\%}$} & \textbf{99.8\%} \textcolor{gray}{$\scriptstyle _{-0.1\%}^{+0.1\%}$} & \textbf{100.0\%} \textcolor{gray}{$\scriptstyle _{-0.0\%}^{+0.0\%}$} \\
        RelApprox-Rslm4           & 4+s & \textbf{100.0\%} \textcolor{gray}{$\scriptstyle _{-0.0\%}^{+0.0\%}$} & \textbf{100.0\%} \textcolor{gray}{$\scriptstyle _{-0.0\%}^{+0.0\%}$} & 99.0\% \textcolor{gray}{$\scriptstyle _{-0.3\%}^{+0.2\%}$} & \textbf{99.8\%} \textcolor{gray}{$\scriptstyle _{-0.1\%}^{+0.1\%}$} & \textbf{100.0\%} \textcolor{gray}{$\scriptstyle _{-0.0\%}^{+0.0\%}$} \\
        RelApprox-TurboQuant      & 4+s & \textbf{100.0\%} \textcolor{gray}{$\scriptstyle _{-0.0\%}^{+0.0\%}$} & \textbf{100.0\%} \textcolor{gray}{$\scriptstyle _{-0.0\%}^{+0.0\%}$} & 98.8\% \textcolor{gray}{$\scriptstyle _{-0.3\%}^{+0.3\%}$} & \textbf{99.8\%} \textcolor{gray}{$\scriptstyle _{-0.1\%}^{+0.1\%}$} & \textbf{100.0\%} \textcolor{gray}{$\scriptstyle _{-0.0\%}^{+0.0\%}$} \\
        RelApprox-Slm4            & 4+s & \textbf{100.0\%} \textcolor{gray}{$\scriptstyle _{-0.0\%}^{+0.0\%}$} & \textbf{100.0\%} \textcolor{gray}{$\scriptstyle _{-0.1\%}^{+0.0\%}$} & 97.1\% \textcolor{gray}{$\scriptstyle _{-0.5\%}^{+0.4\%}$} & \textbf{99.5\%} \textcolor{gray}{$\scriptstyle _{-0.1\%}^{+0.1\%}$} & 98.2\% \textcolor{gray}{$\scriptstyle _{-0.2\%}^{+0.3\%}$} \\
        RelApprox-Rslm4Lite       & 4   & \textbf{100.0\%} \textcolor{gray}{$\scriptstyle _{-0.0\%}^{+0.0\%}$} & \textbf{100.0\%} \textcolor{gray}{$\scriptstyle _{-0.0\%}^{+0.0\%}$} & 98.1\% \textcolor{gray}{$\scriptstyle _{-0.4\%}^{+0.4\%}$} & \textbf{99.6\%} \textcolor{gray}{$\scriptstyle _{-0.1\%}^{+0.1\%}$} & \textbf{100.0\%} \textcolor{gray}{$\scriptstyle _{-0.0\%}^{+0.0\%}$} \\
        RelApprox-SSQ4Lite        & 4   & \textbf{100.0\%} \textcolor{gray}{$\scriptstyle _{-0.0\%}^{+0.0\%}$} & \textbf{99.9\%} \textcolor{gray}{$\scriptstyle _{-0.1\%}^{+0.1\%}$} & 94.8\% \textcolor{gray}{$\scriptstyle _{-0.6\%}^{+0.5\%}$} & \textbf{99.2\%} \textcolor{gray}{$\scriptstyle _{-0.2\%}^{+0.2\%}$} & \textbf{99.9\%} \textcolor{gray}{$\scriptstyle _{-0.1\%}^{+0.1\%}$} \\
        RelApproxGlobal-Rslm3     & 3+s & \textbf{100.0\%} \textcolor{gray}{$\scriptstyle _{-0.0\%}^{+0.0\%}$} & \textbf{99.6\%} \textcolor{gray}{$\scriptstyle _{-0.1\%}^{+0.1\%}$} & \textbf{99.6\%} \textcolor{gray}{$\scriptstyle _{-0.1\%}^{+0.1\%}$} & 98.9\% \textcolor{gray}{$\scriptstyle _{-0.2\%}^{+0.2\%}$} & \textbf{100.0\%} \textcolor{gray}{$\scriptstyle _{-0.0\%}^{+0.0\%}$} \\
        RelApprox-Rslm3           & 3+s & \textbf{100.0\%} \textcolor{gray}{$\scriptstyle _{-0.0\%}^{+0.0\%}$} & \textbf{99.4\%} \textcolor{gray}{$\scriptstyle _{-0.1\%}^{+0.2\%}$} & 94.0\% \textcolor{gray}{$\scriptstyle _{-0.8\%}^{+0.6\%}$} & 98.2\% \textcolor{gray}{$\scriptstyle _{-0.3\%}^{+0.3\%}$} & \textbf{99.7\%} \textcolor{gray}{$\scriptstyle _{-0.1\%}^{+0.1\%}$} \\
        RelApproxGlobal-Rslm2     & 2+s & \textbf{100.0\%} \textcolor{gray}{$\scriptstyle _{-0.0\%}^{+0.0\%}$} & 97.2\% \textcolor{gray}{$\scriptstyle _{-0.3\%}^{+0.3\%}$} & 97.2\% \textcolor{gray}{$\scriptstyle _{-0.4\%}^{+0.3\%}$} & 94.2\% \textcolor{gray}{$\scriptstyle _{-0.6\%}^{+0.6\%}$} & \textbf{99.7\%} \textcolor{gray}{$\scriptstyle _{-0.1\%}^{+0.1\%}$} \\
        RelApprox-Rslm2           & 2+s & \textbf{100.0\%} \textcolor{gray}{$\scriptstyle _{-0.0\%}^{+0.0\%}$} & 96.5\% \textcolor{gray}{$\scriptstyle _{-0.4\%}^{+0.4\%}$} & 80.9\% \textcolor{gray}{$\scriptstyle _{-1.1\%}^{+1.0\%}$} & 91.7\% \textcolor{gray}{$\scriptstyle _{-0.7\%}^{+0.6\%}$} & 97.4\% \textcolor{gray}{$\scriptstyle _{-0.3\%}^{+0.3\%}$} \\
        RelApproxGlobal-Rslm1     & 1+s & \textbf{99.5\%} \textcolor{gray}{$\scriptstyle _{-0.2\%}^{+0.2\%}$} & 86.9\% \textcolor{gray}{$\scriptstyle _{-0.7\%}^{+0.8\%}$} & 87.7\% \textcolor{gray}{$\scriptstyle _{-0.8\%}^{+0.8\%}$} & 82.1\% \textcolor{gray}{$\scriptstyle _{-1.0\%}^{+1.1\%}$} & 97.3\% \textcolor{gray}{$\scriptstyle _{-0.4\%}^{+0.3\%}$} \\
        RelApprox-Rslm1           & 1+s & \textbf{99.2\%} \textcolor{gray}{$\scriptstyle _{-0.2\%}^{+0.2\%}$} & 85.1\% \textcolor{gray}{$\scriptstyle _{-0.8\%}^{+0.8\%}$} & 57.6\% \textcolor{gray}{$\scriptstyle _{-1.1\%}^{+1.1\%}$} & 76.0\% \textcolor{gray}{$\scriptstyle _{-1.2\%}^{+1.2\%}$} & 88.6\% \textcolor{gray}{$\scriptstyle _{-0.7\%}^{+0.7\%}$} \\
        \bottomrule
    \end{tabular}

    \vspace{0.5em}
    \small{\textit{Note}: `+s' indicates that one or two extra scaling factors are stored per vector.}
\end{table*}

\begin{table*}[ht] 
    \centering
    \caption{Recall@20@40 Evaluation Results for Selected Codecs.}
    \label{tab:recall_eval_relative2040}
    \renewcommand{\arraystretch}{1.2}
    \begin{tabular}{@{} l c c c c c c @{}}
        \toprule
        \textbf{Codec} & \textbf{Bits/Dim} & \textbf{\texttt{wiki\_full}} & \textbf{\texttt{wiki\_pca}} & \textbf{\texttt{bigann}} & \textbf{\texttt{glove}} & \textbf{\texttt{openai}} \\
        & & ($D=3072$) & ($D=384$) & ($D=128$) & ($D=100$) & ($D=1536$) \\
        \midrule
        RelApproxGlobal-Rslm2 & 2+s & \textbf{100.0\%} \textcolor{gray}{$\scriptstyle _{-0.0\%}^{+0.0\%}$} & \textbf{99.4\%} \textcolor{gray}{$\scriptstyle _{-0.2\%}^{+0.2\%}$} & \textbf{99.3\%} \textcolor{gray}{$\scriptstyle _{-0.2\%}^{+0.2\%}$} & 97.7\% \textcolor{gray}{$\scriptstyle _{-0.4\%}^{+0.3\%}$} & \textbf{100.0\%} \textcolor{gray}{$\scriptstyle _{-0.0\%}^{+0.0\%}$} \\
        RelApproxGlobal-Rslm1 & 1+s & \textbf{100.0\%} \textcolor{gray}{$\scriptstyle _{-0.0\%}^{+0.0\%}$} & 93.1\% \textcolor{gray}{$\scriptstyle _{-0.5\%}^{+0.5\%}$} & 93.5\% \textcolor{gray}{$\scriptstyle _{-0.6\%}^{+0.5\%}$} & 88.2\% \textcolor{gray}{$\scriptstyle _{-0.9\%}^{+0.9\%}$} & \textbf{99.4\%} \textcolor{gray}{$\scriptstyle _{-0.2\%}^{+0.2\%}$} \\
        \bottomrule
    \end{tabular}

    \vspace{0.5em}
    \small{\textit{Note}: `+s' indicates that one or two extra scaling factors are stored per vector.}
\end{table*}

For relative quantization (see Figure \ref{fig:resid}), we precompute approximate vectors using a multi-stage residual quantization pipeline. 
First, we partition the datasets such that there are approximately 100 vectors per partition\footnote{For the large \texttt{bigann} dataset, this means using 2-level hierarchical partitioning tree.}, and assign each vector to its nearest coarse partition center. The resulting first-stage residual is then quantized using ScaNN's Asymmetric Hashing method (PQ) \cite{guo2020scann} with 1 bit per dimension (see Section \ref{sec:e2e} for ScaNN codecs description and note that we could also apply training-free codec for approximate scoring as we do in Section \ref{sec:e2e}). 

We reconstruct the final approximate document vector by summing the assigned partition center and the decoded residual. This reconstructed vector serves as the approximation for RelApprox variants, which subsequently quantize the remaining secondary residual.
The following observations are based on Tables \ref{tab:recall_eval_relative} and \ref{tab:recall_eval_relative2040}, which show the recall for quantized residual vectors used in candidate rescoring:

\begin{enumerate}
    \item \textbf{Superiority to non-relativized encodings}: Across all datasets and bit rates, having the additional information (approximate vector to which we quantize relatively) drastically reduces recall loss. 
    Our relativized 4-bit \rslm encodings achieve 100\% recall, and 3-bit \rslm encodings achieve $\ge$98.9\% recall across all datasets, offering substantial efficiency gains for the current state-of-the-art ANN systems.    
    \item \textbf{Extreme bandwidth efficiency for RAG systems}: In production vector databases, where downstream ranking systems ingest a larger candidate pool, it makes more sense to also look at metrics like Recall@20@40. We observe that our relativized 2-bit Rslm encoding achieves 99.3\%+ recall@20@40 across four out of five datasets (only for the lowest 100 dimensional \texttt{glove} it is 97.7\%); For high-dimensional datasets, our relativized 1-bit Rslm encoding achieves 100\% recall@20@40 on 3072-dim wiki\_full and 99.4\% recall@20@40 on 1536-dim openai. This enables 4x to 8x reduction in rescoring vectors DRAM footprint and bandwidth.
    \item \textbf{Fixing the norm is crucial}: RelApproxGlobal significantly outperforms RelApprox encodings, experimentally confirming that fixing the norm of the full reconstructed vector (approximate + residual) is critical for MIPS ranking.
\end{enumerate}

\subsection{Other Recall Metrics}
We provide additional recall metrics (defined in Section \ref{sec:quality-metrics}) in Table~\ref{tab:all_recall_metrics}. They trend as expected, with the exception that the \texttt{bigann} dataset achieves high 1\% Tolerant recall@20, even though recall@20@X metrics are worse. Looking closer, this is explained by \texttt{bigann} having clusters of documents with very similar vectors (magnified by discretized per-dimension values -- integers between 0 and 255).

\begin{table}[ht]
\centering
\caption{Various recall metrics comparison across selected datasets and codecs.}
\label{tab:all_recall_metrics}
\begin{tabular}{llccc}
\toprule
\textbf{Codec} & \textbf{Metric} & \textbf{\texttt{openai}} & \textbf{\texttt{glove}} & \textbf{\texttt{bigann}} \\
\midrule
\multirow{4}{*}{SQ8} 
 & R@20@20   & 91.4\%          & 98.0\%          & 92.4\%          \\
 & R@20@30   & \textbf{99.6\%} & \textbf{100.0\%} & 98.8\%          \\
 & R@20@40   & \textbf{100.0\%} & \textbf{100.0\%} & \textbf{99.4\%} \\
 & Tolerant  & \textbf{100.0\%} & \textbf{100.0\%} & \textbf{100.0\%} \\
\midrule
\multirow{4}{*}{SSQ4Lite} 
 & R@20@20   & 49.9\%          & 81.0\%          & 49.4\%          \\
 & R@20@30   & 60.6\%          & 93.3\%          & 60.2\%          \\
 & R@20@40   & 68.0\%          & 97.3\%          & 67.6\%          \\
 & Tolerant  & 82.5\%          & 87.4\%          & 81.1\%          \\
\midrule
\multirow{4}{*}{Rslm4} 
 & R@20@20   & 90.5\%          & 86.8\%          & 79.6\%          \\
 & R@20@30   & 98.9\%          & 97.3\%          & 92.3\%          \\
 & R@20@40   & \textbf{99.9\%} & \textbf{99.3\%} & 96.6\%          \\
 & Tolerant  & \textbf{100.0\%} & 92.7\%          & \textbf{99.6\%} \\
\midrule
\multirow{4}{*}{\makecell{RelApproxGlobal- \\ Rslm4}} 
 & R@20@20   & 97.5\%          & 94.9\%          & 95.5\%          \\
 & R@20@30   & \textbf{100.0\%} & \textbf{100.0\%} & \textbf{100.0\%} \\
 & R@20@40   & \textbf{100.0\%} & \textbf{100.0\%} & \textbf{100.0\%} \\
 & Tolerant  & \textbf{100.0\%} & \textbf{99.3\%} & \textbf{100.0\%} \\
\midrule
\multirow{4}{*}{\makecell{RelApproxGlobal- \\ Rslm2}} 
 & R@20@20   & 92.9\%          & 82.9\%          & 86.4\%          \\
 & R@20@30   & \textbf{99.7\%} & 94.2\%          & 97.2\%          \\
 & R@20@40   & \textbf{100.0\%} & 97.7\%          & 99.3\%          \\
 & Tolerant  & \textbf{100.0\%} & 89.0\%          & \textbf{99.8\%} \\
\bottomrule
\end{tabular}
\end{table}

\subsection{Performance Impact and Memory Savings}
\label{subsec:performance}
We measure dot-product throughput and codec operation throughput using a standardized microbenchmark suite on both AMD Milan (x86\_64 AVX2) and Intel Cascade Lake (x86\_64 AVX-512) server platforms to capture different capacity and utilization of L1/L2/L3 CPU caches by those platforms. 

To achieve peak speed, vector quantization algorithms (like \rslm) perform table lookups directly inside CPU registers using byte-shuffle instructions (\texttt{VPSHUFB}). However, because AVX2 hardware restricts these shuffles to 16-byte lanes, codebooks cannot exceed 16 elements without suffering severe performance penalties~\cite{andre2015cache}.

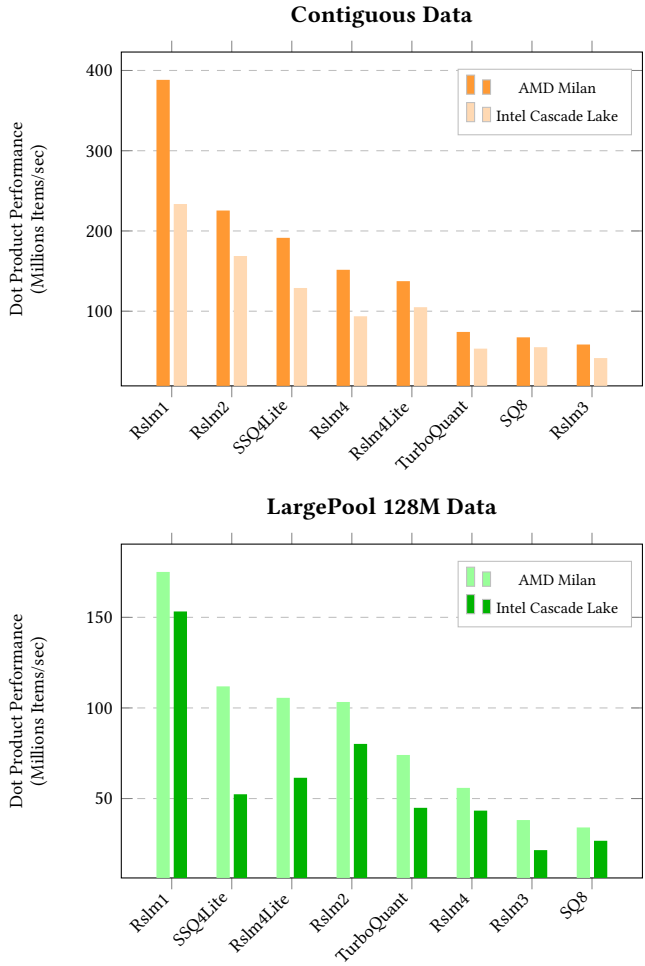
\begin{figure}[htbp]
    \centering
    
    \begin{tikzpicture}
        \begin{axis}[
            title={\textbf{Contiguous Data}},
            ybar=1.5pt, 
            width=\linewidth, 
            height=6cm,       
            bar width=5pt,    
            enlarge x limits=0.12, 
            ylabel={Dot Product Performance \\ (Millions Items/sec)},
            ylabel style={align=center, font=\footnotesize}, 
            tick label style={font=\footnotesize},           
            symbolic x coords={Rslm1, Rslm2, SSQ4Lite, Rslm4, Rslm4Lite, TurboQuant, SQ8, Rslm3},
            xtick=data,
            x tick label style={rotate=45, anchor=east},
            legend style={font=\scriptsize, at={(0.98,0.95)}, anchor=north east, draw=gray!50}, 
            ymajorgrids=true,
            grid style=dashed
        ]

        \addplot[fill=orange!80, draw=none] coordinates {
            (Rslm1, 388.31)
            (Rslm2, 225.47)
            (SSQ4Lite, 191.44)
            (Rslm4, 151.61)
            (Rslm4Lite, 137.43)
            (TurboQuant, 74.16)
            (SQ8, 67.43)
            (Rslm3, 58.49)
        };

        \addplot[fill=orange!30, draw=none] coordinates {
            (Rslm1, 233.57)
            (Rslm2, 168.78)
            (SSQ4Lite, 128.94)
            (Rslm4, 93.68)
            (Rslm4Lite, 104.98)
            (TurboQuant, 53.34)
            (SQ8, 55.15)
            (Rslm3, 41.60)
        };

        \legend{AMD Milan, Intel Cascade Lake}
        \end{axis}
    \end{tikzpicture}
    
    \vspace{0cm} 
    
    \begin{tikzpicture}
        \begin{axis}[
            title={\textbf{LargePool 128M Data}},
            ybar=1.5pt,
            width=\linewidth, 
            height=6cm,
            bar width=5pt,
            enlarge x limits=0.12,
            ylabel={Dot Product Performance \\ (Millions Items/sec)},
            ylabel style={align=center, font=\footnotesize},
            tick label style={font=\footnotesize},
            symbolic x coords={Rslm1, SSQ4Lite, Rslm4Lite, Rslm2, TurboQuant, Rslm4, Rslm3, SQ8},
            xtick=data,
            x tick label style={rotate=45, anchor=east},
            legend style={font=\scriptsize, at={(0.98,0.95)}, anchor=north east, draw=gray!50},
            ymajorgrids=true,
            grid style=dashed
        ]

        \addplot[fill=green!40, draw=none] coordinates {
            (Rslm1, 175.06)
            (SSQ4Lite, 111.90)
            (Rslm4Lite, 105.58)
            (Rslm2, 103.29)
            (TurboQuant, 74.05)
            (Rslm4, 55.88)
            (Rslm3, 38.13)
            (SQ8, 34.04)
        };

        \addplot[fill=green!70!black, draw=none] coordinates {
            (Rslm1, 153.24)
            (SSQ4Lite, 52.36)
            (Rslm4Lite, 61.46)
            (Rslm2, 80.15)
            (TurboQuant, 44.88)
            (Rslm4, 43.33)
            (Rslm3, 21.53)
            (SQ8, 26.69)
        };

        \legend{AMD Milan, Intel Cascade Lake}
        \end{axis}
    \end{tikzpicture}
    
    \caption{Dot product performance on random 256 dimensional embeddings on AMD Milan and Intel Cascade Lake.}
    \label{fig:dot_product_performance}
\end{figure}

The main focus of our benchmarks is OneToMany Dot-product operation, i.e., computing the dot-product of one query vector with many document vectors. Figure \ref{fig:dot_product_performance} shows the throughput if the scored vectors are contiguous in memory (Plot 1) or whether they are randomly spread within the large pool of the data (Plot 2).
The latter is modeling the real-world rescoring use-case better, while the former shows the headroom if DRAM bandwidth was not an issue.
The plots clearly show:
\begin{itemize}
    \item A smaller RAM footprint not only saves cost (as less RAM is needed), but it also leads, because of higher throughput, to much better end-to-end system performance.
    \item The cacheline-optimized Rslm4Lite significantly outperforms Rslm4 on the LargePool benchmark.
    \item Even though Rslm4Lite uses the fixed Lloyd-Max codebook, in contrast to SSQ4Lite not having any codebook, it can compete on the throughput thanks to highly optimized SIMD CPU instructions.
\end{itemize}

We also evaluated the throughput of Compress, Decompress, and \allowbreak\mbox{Precompute-Query} operations. Compress is applied on each vector once at index building time, making it less critical, but it can still be a substantial cost. For codecs utilizing rotations, it makes a big difference whether it is full random rotation, FWHT on the whole vector, or very efficient per-block approach we apply in Rslm codecs family. This rotation needs to be applied also at serving time, once per query, so that we can then directly compute dot-products in the rotated space instead of decompressing all the vectors on the fly. Being able to have the fast rotation, whatever the vector dimensionality is, was one of the reasons that, even though Rslm4 is similar to 4-bit TurboQuant, we created an Rslm4 version (see Table \ref{tab:quantizers_methods}). 
In production, non-standard dimensionalities arise (e.g., $D=2049$, where an extra dimension encodes domain-specific metadata). While performing comparably at $D=256$, TurboQuant experiences significant slowdown during Compress and Precompute-Query at $D=2049$ due to its reliance on full-dimensional rotations -- on AMD Milan Turboquant can compress 3.5k items per second, while Rslm4 compresses 128k items per second. 

%% file: sections/results-end-to-end-1908.tex
\subsection{End-to-End Comparison with Faiss and ScaNN} \label{sec:e2e}

In this section, we perform an end-to-end comparison of the Rslm codecs with the quantization methods available in the Faiss\footnote{\url{https://github.com/facebookresearch/faiss}} and ScaNN\footnote{\url{https://github.com/google-research/google-research/tree/master/scann}} libraries on the full \texttt{glove} and \texttt{openai} datasets over 500 queries. The goal of this comparison is to evaluate how these quantization methods perform in realistic vector search scenarios. In contrast to previous subsections, where we focused on rescoring phase (i.e., comparing encodings of vectors relative to approximate vectors), here we show the benefits also for approximate scoring (i.e., comparing encodings relative to partition centers).

\subsubsection{Setup}
The evaluation pipeline consists of three phases:
\begin{enumerate}
    \item \textbf{Dataset Partitioning}: The datasets are clustered to define coarse search partitions.
    \item \textbf{Residual Quantization}: We quantize the residual vectors (the difference between each vector and its assigned partition centroid) using the respective codecs.
    \item \textbf{Query Routing and Scanning}: At query time, we identify the $n_{\text{probe}}$ closest partitions (defining the search budget) based on the dot product to the centroids. We then scan the quantized residual vectors within these active partitions to retrieve the top-$k$ nearest neighbors.
\end{enumerate}

For the partitioning step, we configure both Faiss (using an IVF4096 Flat index) and ScaNN (using a single-level \kmeans tree) to partition the dataset into 4,096 partitions for \texttt{glove} and 8,192 for \texttt{openai}.\footnote{This corresponds to $4 \times \sqrt{N}$ rounded up to the next power of 2, where $N$ is the size of the considered dataset.} To ensure a fair comparison, both partitioners are trained using identical parameters: $25$ \kmeans iterations, a training sample size of 1,048,576 vectors for \texttt{glove} and 2,097,152 for \texttt{openai} (corresponding to 256 points per centroid on average), and random initialization. 
To evaluate each baseline under its optimal settings, Faiss and ScaNN codecs are run on their respective default native partitions (spherical \kmeans for Faiss, and standard Euclidean \kmeans for ScaNN), while our proposed codecs are evaluated directly on ScaNN's partitions.

\paragraph{Faiss Codecs} \label{sec:faiss-codecs}
We briefly describe the Faiss codecs which we use in our evaluation. For more details, please see \cite{douze2025faiss}.
\begin{itemize}
    \item \textbf{Scalar Quantization (Faiss SQ)}: Maps floating-point values to equal-sized bins within a dynamically determined range. We evaluate both the default \textit{non-uniform} version, which maintains an independent range per dimension (e.g., {Faiss SQ4}),
    and the \textit{uniform} version, which shares a single range across all dimensions (e.g., {FaissSQ4 Uniform}). 
    
    \item \textbf{Product Quantization ({Faiss PQ})}: Decomposes the vector into $M$ sub-vectors of dimension $D/M$ and quantizes each subspace independently using a codebook trained via \kmeans. For instance, {FaissPQ50x4} partitions the 100-dimensional GloVe vector into 50 2-dimensional sub-vectors, quantizing each with a 4-bit ($K=16$ centroids) codebook to yield an overall rate of 2 bits/dimension.
\end{itemize}

\paragraph{ScaNN Codecs}

From the ScaNN library, we evaluate Asymmetric Hashing (AH), which is ScaNN's implementation of trained PQ.
While the codebook itself is trained using standard \kmeans,
the vectors can be encoded based on either MSE or anisotropic loss \cite{guo2020scann}. Unlike MSE, anisotropic loss prioritizes MIPS ranking by penalizing parallel error, i.e., the quantization error component along the vector's direction that directly alters inner product values. A threshold $t \in [0,1)$ specifies the expected similarity regime: a higher $t$ assumes closer nearest neighbors and penalizes parallel error more aggressively, whereas a lower $t$ converges toward MSE.


Faiss and ScaNN also provide the option to apply a random orthogonal projection matrix to rotate the vectors before quantization. This distributes coordinate variance uniformly, similar in goal to Rslm's FWHT-based pipeline, but requires dense $O(D^2)$ matrix multiplication. We denote by \textit{Faiss Rotated} and \textit{ScaNN Rotated} the codecs for which such a rotation was applied prior to quantization.

\paragraph{Results}

\begin{figure*}[ht]
    \centering
    \begin{subfigure}{0.47\textwidth}
         \includegraphics[width=\linewidth]{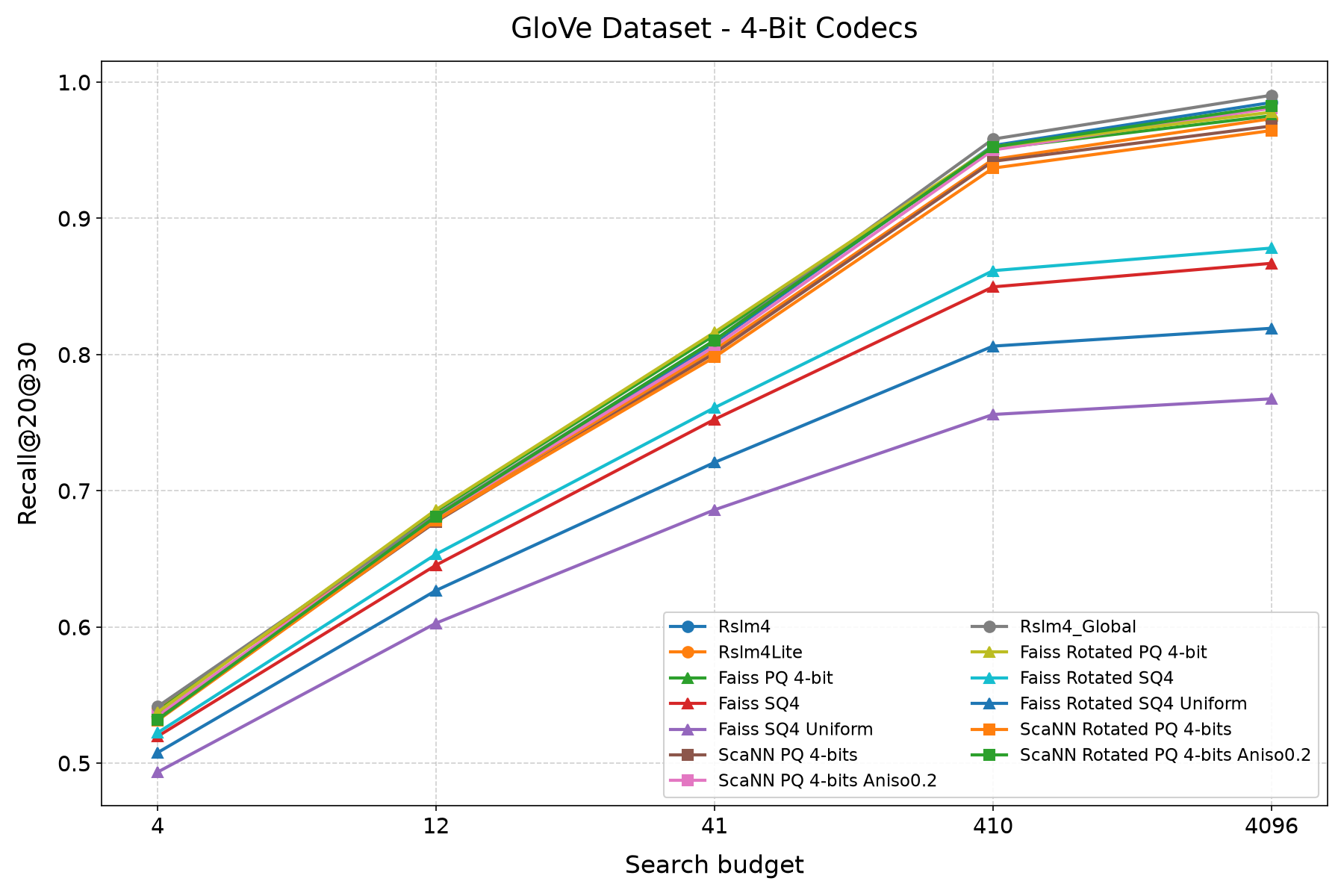}
        \label{fig:4bit-glove}
    \end{subfigure}\hfill
    \begin{subfigure}{0.47\textwidth}
        \includegraphics[width=\linewidth]{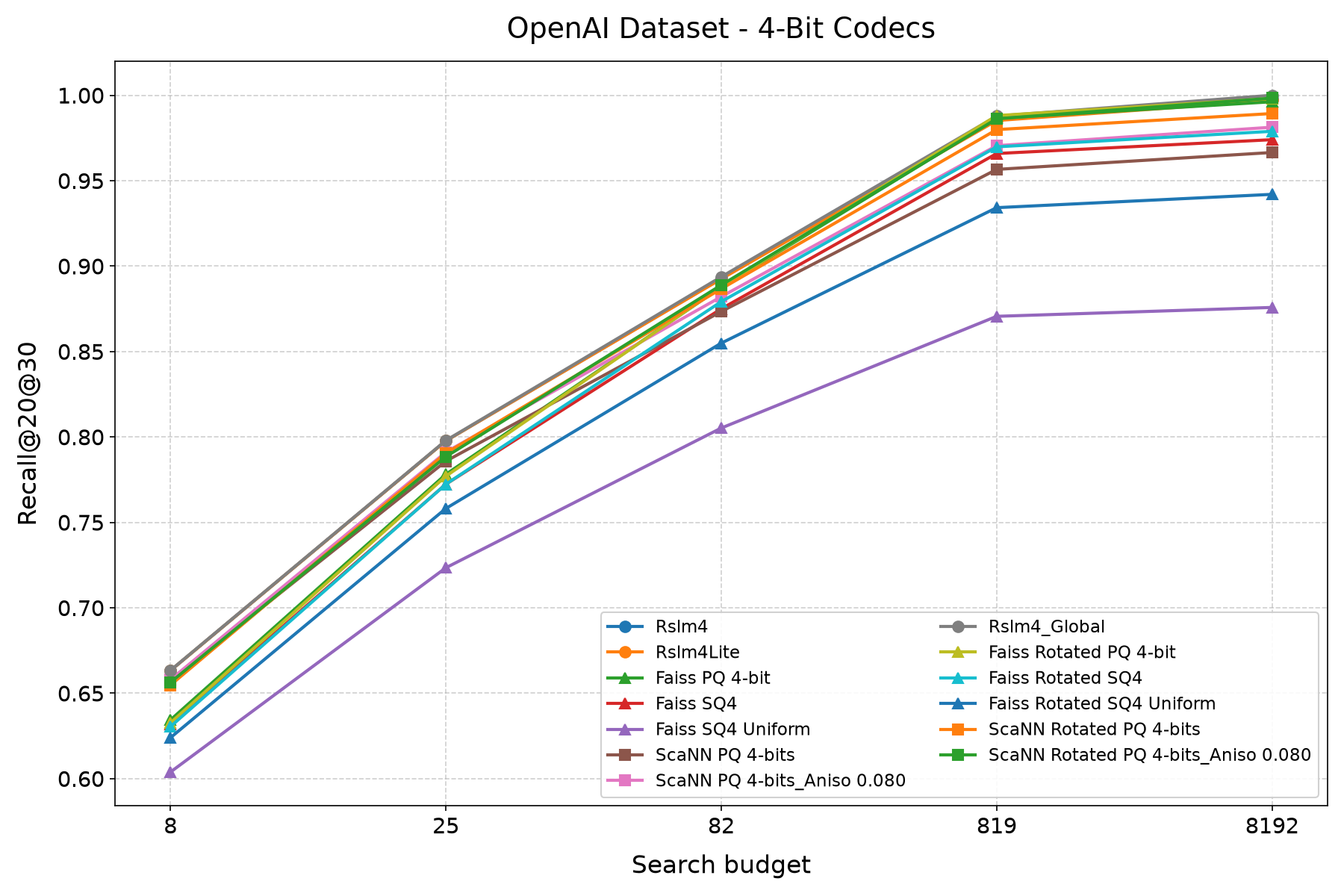}
        \label{fig:4bit-openai}
    \end{subfigure}
    
    \vspace{-3mm} 
    
    \begin{subfigure}{0.47\textwidth}
        \includegraphics[width=\linewidth]{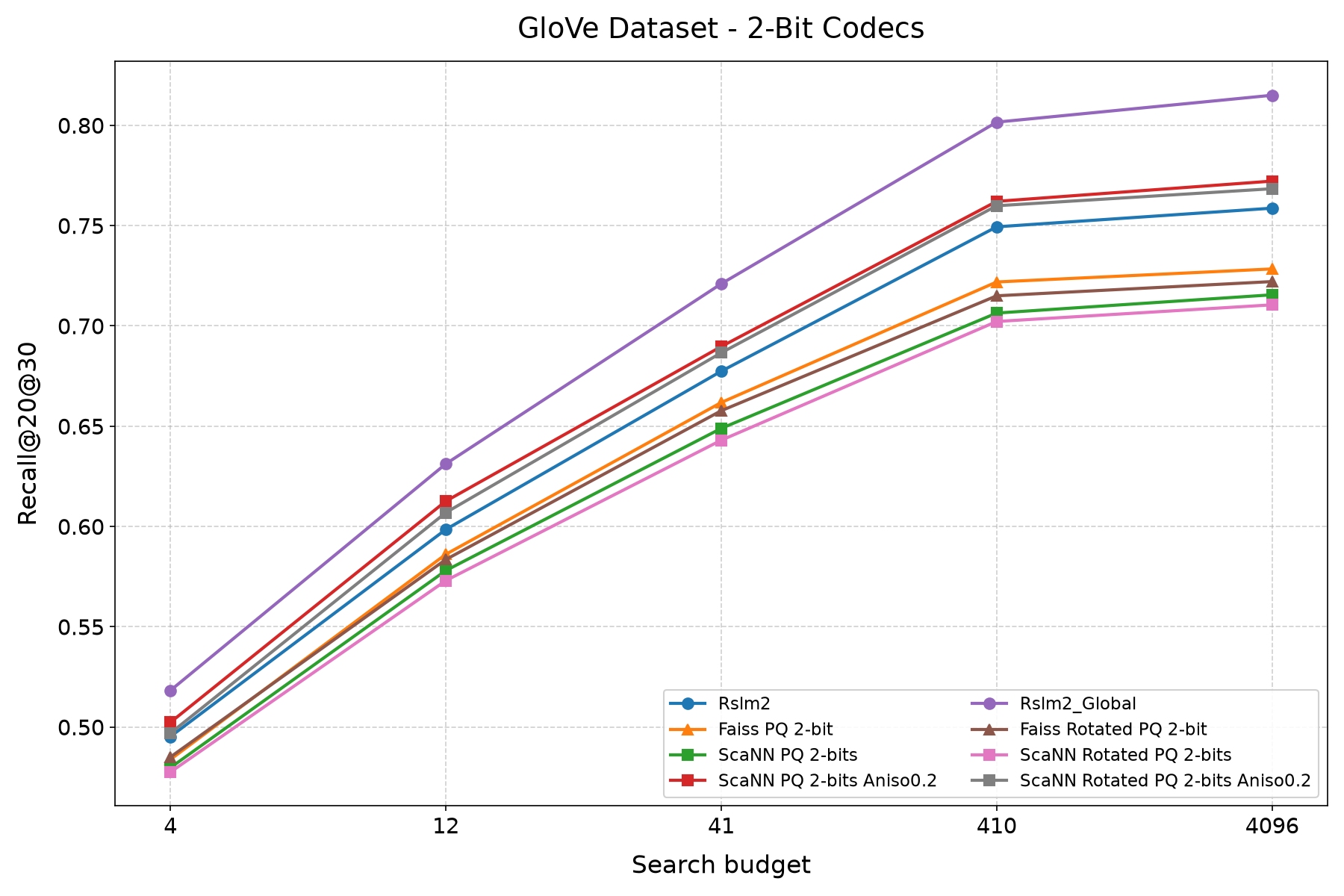}
        \label{fig:2bit-glove}
    \end{subfigure}\hfill
    \begin{subfigure}{0.47\textwidth}
        \includegraphics[width=\linewidth]{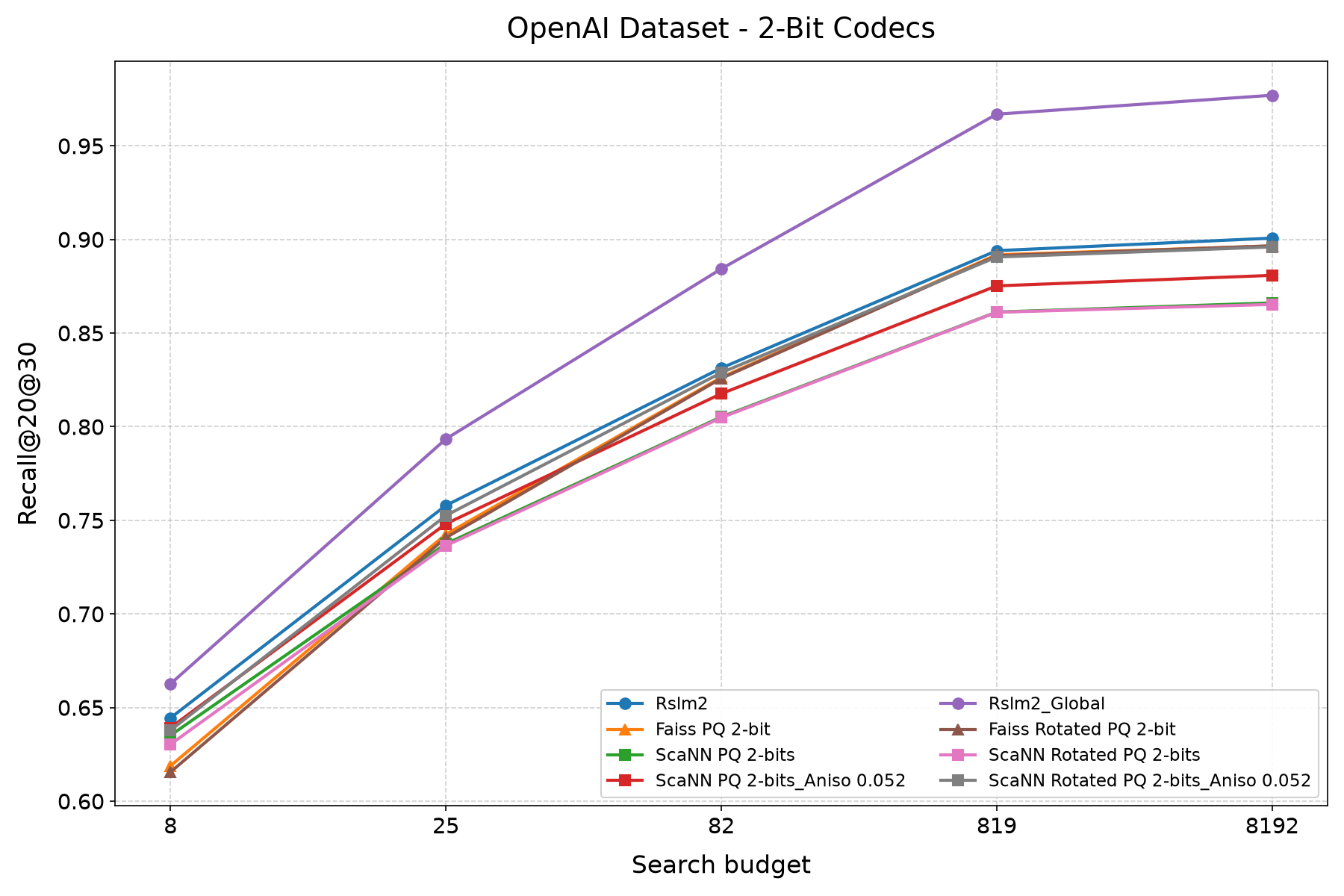}
        \label{fig:2bit-openai}
    \end{subfigure}

    \vspace{-3mm} 

    \begin{subfigure}{0.47\textwidth}
        \includegraphics[width=\linewidth]{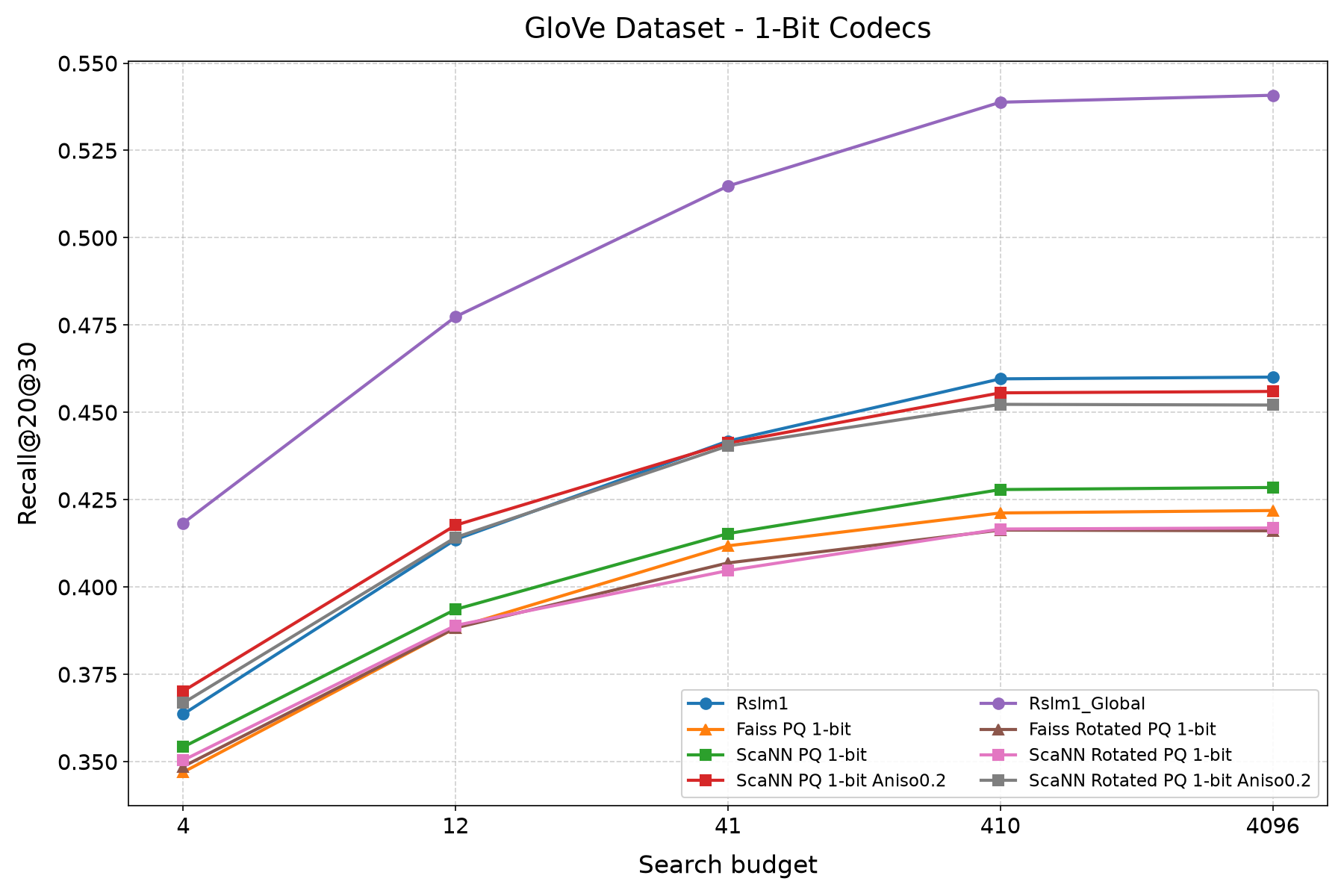}
        \label{fig:1bit-glove}
    \end{subfigure}\hfill
    \begin{subfigure}{0.47\textwidth}
        \includegraphics[width=\linewidth]{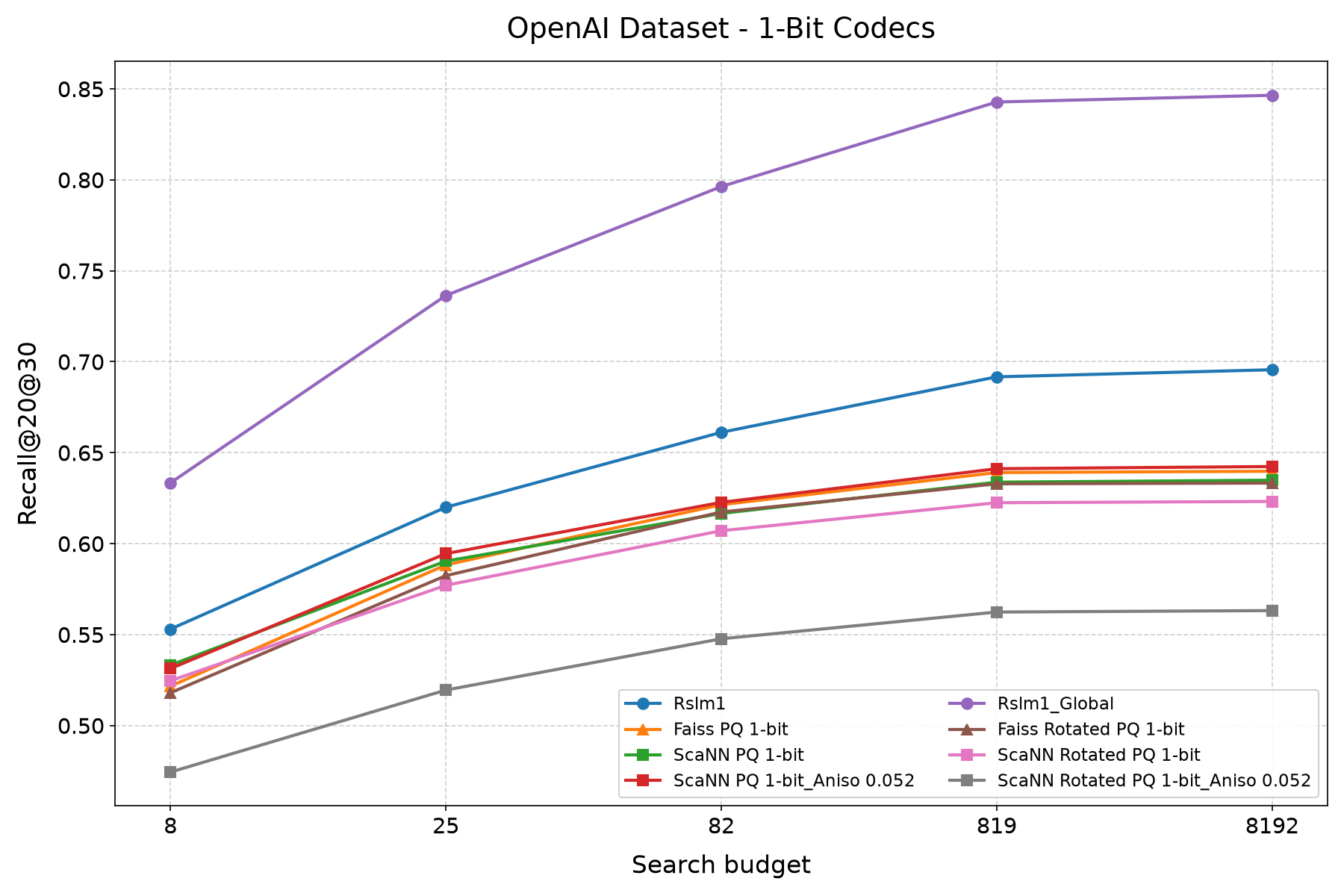}
        \label{fig:1bit-openai}
    \end{subfigure}

    \caption{Recall@20@30 results for Faiss, ScaNN, and Rslm codecs across different search budgets.}
    \label{fig:all-codecs-comparison}
\end{figure*}

We analyze recall as a function of the search budget $n_{\text{probe}}$, testing 0.1\%, 0.3\%, 1\%, 10\%, and 100\% of the total partitions (i.e., $n_{\text{probe}} \in \{4, 12, 41, 410, 4096\}$ for  \texttt{glove} and $n_{\text{probe}} \in \{8, 25, 82, 819, 8192\}$ for \texttt{openai}).
Figure \ref{fig:all-codecs-comparison} illustrates recall@20@30 obtained for different search budgets with 4- to 1-bit codecs. 

ScaNN results obtained with anisotropic vector quantization (AVQ) using threshold $t$ are marked with the suffix ``Aniso$t$''. 
In our experiments, we evaluated the thresholds $t \in \{0.1, 0.2, 0.3\}$ for \texttt{glove}, and included in the plots the value that led to the best recall. Note that  using the same thresholds for \texttt{openai}, ScaNN AVQ led to considerably worse results than standard ScaNN PQ at 2- and especially 1-bit levels (e.g., for $n_{\text{probe}}=82$ and 1-bit level, ScaNN PQ Aniso0.2 has 18.7\% recall@20@30, while ScaNN PQ has 61.66\% recall@20@30). This is because in the anisotropic loss function, the parameter  that is used to penalize the parallel error depends on both the threshold and the vector dimension.\footnote{Specifically, the parameter is $\eta = \frac{t^2 (D-1)}{1-t^2}$ and the anisotropic loss is $\mathcal{L} = \eta \Vert{}e_{\parallel}\Vert{}^2 + \Vert{}e_{\perp}\Vert{}^2$, where $e_{\parallel}$ and $e_{\perp}$ denote the parallel and perpendicular errors, respectively.} Thus, configuring $t$ in ScaNN AVQ requires tuning and depends on the dataset. We evaluated the equivalent thresholds for \texttt{openai} that imply the same parallel multiplier as the thresholds used for \texttt{glove} and included in the plots the best one.
In what follows, we discuss the \texttt{glove} results, noting that the \texttt{openai} results exhibit similar behavior.

In an end-to-end ANN search pipeline, retrieval quality is determined by both coarse partitioning errors (missing the correct cluster) and quantization errors (distance distortion during scoring). The relative impact of these two error sources depends  on the codec's bitrate. 
At higher bitrates (such as 4-bits), quantization error is negligible (see Table \ref{tab:recall_eval_non_relative}), meaning retrieval quality depends primarily on partitioning errors. In this regime, increasing the search budget yields substantial recall gains (e.g., from $54.1\%$ at $n_{\text{probe}}=4$ to $95.8\%$ at $n_{\text{probe}}=410$ for {Rslm4\_Global}). 
Conversely, at lower bitrates, quantization noise becomes the primary bottleneck, capping the maximum achievable recall even when searching the entire database ($54.0\%$ recall for {Rslm1\_Global} at $n_{\text{probe}}=4096$). 

\paragraph{Similar Recall Without Codebook Training}
Unlike standard PQ in Faiss and ScaNN, which requires running \kmeans to train dataset-specific codebooks, Rslm utilizes fixed Lloyd-Max codebooks. This training-free approach is made possible by Rslm's rotation pipeline. Our evaluation across all bit levels shows that Rslm codecs achieve comparable or superior recall to the trained Faiss and ScaNN configurations. By eliminating the \kmeans training phase, Rslm enables fast deployment on new datasets (where we do not have enough samples for training) and considerably reduces indexing complexity.


\paragraph{Low-Bitrate Superiority of Rslm}
At highly compressed rates (1- and 2-bit, which are also the default settings in ScaNN), Rslm\_Global codecs significantly outperform  ScaNN and Faiss. For instance, for $n_{\text{probe}} = 4096$, Rslm1\_Global reaches $54.0\%$ recall, outperforming ScaNN's AVQ (45.6\%) and Faiss's PQ (42.1\%). At 2-bits, Rslm2\_Global outperforms ScaNN's best variant by $4.3\%$ (Faiss by $8.7\%$). 

The superior results of Rslm low-bit codecs can be attributed to fixing the norm of the vector for ranking.
While ScaNN AVQ improves the results compared to simple ScaNN PQ, it requires tuning.
Rslm achieves a similar effect by strictly enforcing norm preservation in the reconstructed vector. Although this requires storing an auxiliary scalar per vector, it avoids complex anisotropic optimization. 
For low-dimensional datasets like \texttt{glove}, this extra scalar increases the memory requirements by 15\% for 1-bit codecs (Faiss PQ requires 13, while Rslm1 requires 15 bytes per vector), while for higher-dimensional datasets like \texttt{openai}, the increase is modest, only 1\% (192 in Faiss PQ vs. 194 in Rslm1). 
Furthermore, globally scaling the combined centroid-plus-residual vector yields an additional increase in recall (this extra scalar means that two scalars are 31\% increase in size for 1-bit codecs for \texttt{glove} while only 2\% increase for high-dimensional \texttt{openai}).
This allows us to achieve similar recall while searching significantly fewer partitions (if this is the final scoring), or rescoring significantly fewer documents (if this is approximate scoring phase).
We acknowledge that storing an extra scalar on low-bit codecs makes it slightly unfair for comparisons on lower dimensional datasets like \texttt{glove} but is negligible for higher-dimensional datasets like \texttt{openai} or \texttt{wiki\_full}. 


\paragraph{Codebook Training Differences}
Finally, we note that while both ScaNN and Faiss implement PQ, they differ in their codebook training initialization. ScaNN utilizes k-means++ initialization, 
whereas Faiss relies on standard random initialization. Consequently, the recall results for these structurally equivalent PQ configurations are close but not identical. For instance, at $n_{\text{probe}} = 41$, 4-bit Faiss PQ slightly outperforms ScaNN PQ ($81.4\%$ recall vs. $80.1\%$). At 2 bits/dimension, the difference in recall is 1.3\% ($66.1\%$ in Faiss, $64.8\%$ in ScaNN), while at 1 bit/dimension, the results are very close ( $41.5\%$ in ScaNN, $41.1\%$ in Faiss).
We also note that applying random orthogonal rotations improves SQ, but it generally does not improve PQ. This happens because random rotations spread variance evenly across dimensions, helping SQ's rigid coordinate ranges avoid clipping high-variance coordinates, whereas PQ already handles multi-dimensional spaces by clustering sub-vectors directly.

%% file: sections/conclusions.tex
\section{Conclusions}
We introduced Rslm, a family of training-free vector quantization codecs spanning 1 to 4 bits per dimension, designed to save memory resources and eliminate memory-bandwidth bottlenecks in large-scale ANN search. We optimized them for encoding the residual vectors and demonstrated that the current state-of-the-art systems could benefit from 2$\times$ to 4$\times$ improvements. One can take these as direct savings. Another option is to fund with them other optimizations like SOAR \cite{sun2023soar} which needs extra memory but leads to a much smaller number of partitions that need to be searched.

Our major contribution is not only to correct the norm of the residual vector, but of the final reconstructed vector. This is a simple and yet highly effective solution to preserve geometric integrity for MIPS, experimentally outperforming complex anisotropic loss formulations at low-bitrate compressions of residual vectors.

We achieved highly scalable, hardware-efficient processing, by focusing on details of the whole end-to-end pipeline: (1) We proposed block-wise cascaded FWHT which effectively approximates dense global rotations with linear-like complexity even for strange input dimensionalities. (2) We kept codebooks of size at most 16 to be able to utilize fast AVX SIMD instructions. (3) We introduced Rslm4Lite codec which hides a 16-bit scale factor to achieve perfect cache-line alignment providing a major throughput advantage for production vector databases.

And finally, our empirical results demonstrate that completely training-free codecs can match or exceed the semantic recall of highly optimized, data-dependent industry standards (Faiss PQ, ScaNN AH) without the burden of offline k-means training. Having training-free codecs both for the approximate scoring phase (as shown in Section \ref{sec:e2e}) and the rescoring phase (as shown in Subsection \ref{sec:results-relative}) of ANN paves the way for simplified and lower-cost production vector database architectures.

%% file: sections/reporducibility.tex
\section{Reproducibility}
We provide a reference Python implementation of our codecs within a Colab notebook which reproduces quality metrics on full \texttt{glove} dataset, both for non-relative and relative quantizations. 

See \url{https://github.com/google-research/google-research/tree/master/rslm}.

%% file: main.bbl

\begin{thebibliography}{54}


\ifx \showCODEN    \undefined \def \showCODEN     #1{\unskip}     \fi
\ifx \showDOI      \undefined \def \showDOI       #1{#1}\fi
\ifx \showISBNx    \undefined \def \showISBNx     #1{\unskip}     \fi
\ifx \showISBNxiii \undefined \def \showISBNxiii  #1{\unskip}     \fi
\ifx \showISSN     \undefined \def \showISSN      #1{\unskip}     \fi
\ifx \showLCCN     \undefined \def \showLCCN      #1{\unskip}     \fi
\ifx \shownote     \undefined \def \shownote      #1{#1}          \fi
\ifx \showarticletitle \undefined \def \showarticletitle #1{#1}   \fi
\ifx \showURL      \undefined \def \showURL       {\relax}        \fi
\providecommand\bibfield[2]{#2}
\providecommand\bibinfo[2]{#2}
\providecommand\natexlab[1]{#1}
\providecommand\showeprint[2][]{arXiv:#2}

\bibitem[\protect\citeauthoryear{Aguerrebere, Bhati, Hildebrand, Tepper, and Willke}{Aguerrebere et~al\mbox{.}}{2023}]%
        {aguerrebere2023similarity}
\bibfield{author}{\bibinfo{person}{Cecilia Aguerrebere}, \bibinfo{person}{Ishwar~Singh Bhati}, \bibinfo{person}{Mark Hildebrand}, \bibinfo{person}{Mariano Tepper}, {and} \bibinfo{person}{Theodore Willke}.} \bibinfo{year}{2023}\natexlab{}.
\newblock \showarticletitle{Similarity Search in the Blink of an Eye with Compressed Indices}.
\newblock \bibinfo{journal}{\emph{Proceedings of the VLDB Endowment}} \bibinfo{volume}{16}, \bibinfo{number}{11} (\bibinfo{year}{2023}), \bibinfo{pages}{3433--3446}.
\newblock
\urldef\tempurl%
\url{https://doi.org/10.14778/3611479.3611537}
\showDOI{\tempurl}


\bibitem[\protect\citeauthoryear{Andr{\'e}, Kermarrec, and Le~Scouarnec}{Andr{\'e} et~al\mbox{.}}{2015}]%
        {andre2015cache}
\bibfield{author}{\bibinfo{person}{Fabien Andr{\'e}}, \bibinfo{person}{Anne-Marie Kermarrec}, {and} \bibinfo{person}{Nicolas Le~Scouarnec}.} \bibinfo{year}{2015}\natexlab{}.
\newblock \showarticletitle{Cache locality is not enough: High-performance nearest neighbor search with product quantization fast scan}.
\newblock \bibinfo{journal}{\emph{Proceedings of the VLDB Endowment}} \bibinfo{volume}{9}, \bibinfo{number}{4} (\bibinfo{year}{2015}), \bibinfo{pages}{288--299}.
\newblock


\bibitem[\protect\citeauthoryear{Ann, Lee, and Oh}{Ann et~al\mbox{.}}{2026}]%
        {ann2026block}
\bibfield{author}{\bibinfo{person}{Heesang Ann}, \bibinfo{person}{Joongkyu Lee}, {and} \bibinfo{person}{Min-hwan Oh}.} \bibinfo{year}{2026}\natexlab{}.
\newblock \showarticletitle{Block-Sphere Vector Quantization}.
\newblock \bibinfo{journal}{\emph{arXiv preprint arXiv:2605.19972}} (\bibinfo{year}{2026}).
\newblock


\bibitem[\protect\citeauthoryear{Ashkboos, Mohtashami, Croci, Li, Cameron, Jaggi, Alistarh, Hoefler, and Hensman}{Ashkboos et~al\mbox{.}}{2024}]%
        {ashkboos2024quarot}
\bibfield{author}{\bibinfo{person}{Saleh Ashkboos}, \bibinfo{person}{Amirkeivan Mohtashami}, \bibinfo{person}{Maximilian~L Croci}, \bibinfo{person}{Bo Li}, \bibinfo{person}{Pashmina Cameron}, \bibinfo{person}{Martin Jaggi}, \bibinfo{person}{Dan Alistarh}, \bibinfo{person}{Torsten Hoefler}, {and} \bibinfo{person}{James Hensman}.} \bibinfo{year}{2024}\natexlab{}.
\newblock \showarticletitle{Quarot: Outlier-free 4-bit inference in rotated llms}.
\newblock \bibinfo{journal}{\emph{Advances in Neural Information Processing Systems}}  \bibinfo{volume}{37} (\bibinfo{year}{2024}), \bibinfo{pages}{100213--100240}.
\newblock


\bibitem[\protect\citeauthoryear{Aum{\"u}ller, Bernhardsson, and Faithfull}{Aum{\"u}ller et~al\mbox{.}}{2020}]%
        {aumuller2020ann}
\bibfield{author}{\bibinfo{person}{Martin Aum{\"u}ller}, \bibinfo{person}{Erik Bernhardsson}, {and} \bibinfo{person}{Alexander Faithfull}.} \bibinfo{year}{2020}\natexlab{}.
\newblock \showarticletitle{ANN-Benchmarks: A benchmarking tool for approximate nearest neighbor algorithms}.
\newblock \bibinfo{journal}{\emph{Information Systems}}  \bibinfo{volume}{87} (\bibinfo{year}{2020}), \bibinfo{pages}{101374}.
\newblock


\bibitem[\protect\citeauthoryear{Auvolat, Chandar, Vincent, Larochelle, and Bengio}{Auvolat et~al\mbox{.}}{2015}]%
        {auvolat2015clusteringefficientapproximatemaximum}
\bibfield{author}{\bibinfo{person}{Alex Auvolat}, \bibinfo{person}{Sarath Chandar}, \bibinfo{person}{Pascal Vincent}, \bibinfo{person}{Hugo Larochelle}, {and} \bibinfo{person}{Yoshua Bengio}.} \bibinfo{year}{2015}\natexlab{}.
\newblock \bibinfo{title}{Clustering is Efficient for Approximate Maximum Inner Product Search}.
\newblock
\newblock
\showeprint[arxiv]{1507.05910}~[cs.LG]
\urldef\tempurl%
\url{https://arxiv.org/abs/1507.05910}
\showURL{%
\tempurl}


\bibitem[\protect\citeauthoryear{Bentley}{Bentley}{1975}]%
        {bentley1975tree}
\bibfield{author}{\bibinfo{person}{Jon~Louis Bentley}.} \bibinfo{year}{1975}\natexlab{}.
\newblock \showarticletitle{Multidimensional binary search trees used for associative searching}.
\newblock \bibinfo{journal}{\emph{Commun. ACM}} \bibinfo{volume}{18}, \bibinfo{number}{9} (\bibinfo{year}{1975}), \bibinfo{pages}{509--517}.
\newblock


\bibitem[\protect\citeauthoryear{Bingham and Mannila}{Bingham and Mannila}{2001}]%
        {bingham2001random}
\bibfield{author}{\bibinfo{person}{Ella Bingham} {and} \bibinfo{person}{Heikki Mannila}.} \bibinfo{year}{2001}\natexlab{}.
\newblock \showarticletitle{Random projection in dimensionality reduction: applications to image and text data}. In \bibinfo{booktitle}{\emph{Proceedings of the seventh ACM SIGKDD international conference on Knowledge discovery and data mining}}. \bibinfo{pages}{245--250}.
\newblock


\bibitem[\protect\citeauthoryear{Bruch}{Bruch}{2024}]%
        {bruch2024vectorretrieval}
\bibfield{author}{\bibinfo{person}{Sebastian Bruch}.} \bibinfo{year}{2024}\natexlab{}.
\newblock \bibinfo{booktitle}{\emph{Foundations of Vector Retrieval}}.
\newblock \bibinfo{publisher}{Springer Nature Switzerland}.
\newblock
\showISBNx{9783031551826}
\urldef\tempurl%
\url{https://doi.org/10.1007/978-3-031-55182-6}
\showDOI{\tempurl}


\bibitem[\protect\citeauthoryear{Chee, Cai, Kuleshov, and De~Sa}{Chee et~al\mbox{.}}{2023}]%
        {chee2023quip}
\bibfield{author}{\bibinfo{person}{Jerry Chee}, \bibinfo{person}{Yaohui Cai}, \bibinfo{person}{Volodymyr Kuleshov}, {and} \bibinfo{person}{Christopher~M De~Sa}.} \bibinfo{year}{2023}\natexlab{}.
\newblock \showarticletitle{Quip: 2-bit quantization of large language models with guarantees}.
\newblock \bibinfo{journal}{\emph{Advances in neural information processing systems}}  \bibinfo{volume}{36} (\bibinfo{year}{2023}), \bibinfo{pages}{4396--4429}.
\newblock


\bibitem[\protect\citeauthoryear{Chen, Zhao, Wang, Li, Liu, Li, Yang, and Wang}{Chen et~al\mbox{.}}{2021}]%
        {chen2021spann}
\bibfield{author}{\bibinfo{person}{Qi Chen}, \bibinfo{person}{Bing Zhao}, \bibinfo{person}{Haidong Wang}, \bibinfo{person}{Mingqin Li}, \bibinfo{person}{Chuanjie Liu}, \bibinfo{person}{Zengzhong Li}, \bibinfo{person}{Mao Yang}, {and} \bibinfo{person}{Jingdong Wang}.} \bibinfo{year}{2021}\natexlab{}.
\newblock \showarticletitle{Spann: Highly-efficient billion-scale approximate nearest neighborhood search}.
\newblock \bibinfo{journal}{\emph{Advances in Neural Information Processing Systems}}  \bibinfo{volume}{34} (\bibinfo{year}{2021}), \bibinfo{pages}{5199--5212}.
\newblock


\bibitem[\protect\citeauthoryear{Chen, Guan, and Wang}{Chen et~al\mbox{.}}{2010}]%
        {chen2010rq}
\bibfield{author}{\bibinfo{person}{Yongjian Chen}, \bibinfo{person}{Tao Guan}, {and} \bibinfo{person}{Cheng Wang}.} \bibinfo{year}{2010}\natexlab{}.
\newblock \showarticletitle{Approximate nearest neighbor search by residual vector quantization}.
\newblock \bibinfo{journal}{\emph{Sensors}} \bibinfo{volume}{10}, \bibinfo{number}{12} (\bibinfo{year}{2010}), \bibinfo{pages}{11259--11273}.
\newblock


\bibitem[\protect\citeauthoryear{Cohan, Dernoncourt, Kim, Bui, Kim, Chang, and Goharian}{Cohan et~al\mbox{.}}{2018}]%
        {Cohan2018}
\bibfield{author}{\bibinfo{person}{Arman Cohan}, \bibinfo{person}{Franck Dernoncourt}, \bibinfo{person}{Doo~Soon Kim}, \bibinfo{person}{Trung Bui}, \bibinfo{person}{Seokhwan Kim}, \bibinfo{person}{Walter Chang}, {and} \bibinfo{person}{Nazli Goharian}.} \bibinfo{year}{2018}\natexlab{}.
\newblock \showarticletitle{A Discourse-Aware Attention Model for Abstractive Summarization of Long Documents}.
\newblock \bibinfo{journal}{\emph{Proceedings of the 2018 Conference of the North American Chapter of the Association for Computational Linguistics: Human Language Technologies, Volume 2 (Short Papers)}} (\bibinfo{year}{2018}).
\newblock
\urldef\tempurl%
\url{https://doi.org/10.18653/v1/n18-2097}
\showDOI{\tempurl}


\bibitem[\protect\citeauthoryear{Coles}{Coles}{2001}]%
        {coles2001introduction}
\bibfield{author}{\bibinfo{person}{Stuart Coles}.} \bibinfo{year}{2001}\natexlab{}.
\newblock \bibinfo{booktitle}{\emph{An Introduction to Statistical Modeling of Extreme Values}}.
\newblock \bibinfo{publisher}{Springer London}, \bibinfo{address}{London}.
\newblock
\showISBNx{978-1-85233-459-8}
\urldef\tempurl%
\url{https://doi.org/10.1007/978-1-4471-3675-0}
\showDOI{\tempurl}


\bibitem[\protect\citeauthoryear{Dai, Yan, Ng, Liu, and Cheng}{Dai et~al\mbox{.}}{2020}]%
        {dai2020norm}
\bibfield{author}{\bibinfo{person}{Xinyan Dai}, \bibinfo{person}{Xiao Yan}, \bibinfo{person}{Kelvin~KW Ng}, \bibinfo{person}{Jie Liu}, {and} \bibinfo{person}{James Cheng}.} \bibinfo{year}{2020}\natexlab{}.
\newblock \showarticletitle{Norm-Explicit Quantization: Improving Vector Quantization for Maximum Inner Product Search}. In \bibinfo{booktitle}{\emph{Proceedings of the AAAI Conference on Artificial Intelligence}}, Vol.~\bibinfo{volume}{34}. \bibinfo{pages}{3702--3709}.
\newblock


\bibitem[\protect\citeauthoryear{Dasgupta and Sinha}{Dasgupta and Sinha}{2013}]%
        {dasgupta2013tree}
\bibfield{author}{\bibinfo{person}{Sanjoy Dasgupta} {and} \bibinfo{person}{Kaushik Sinha}.} \bibinfo{year}{2013}\natexlab{}.
\newblock \showarticletitle{Randomized partition trees for exact nearest neighbor search}. In \bibinfo{booktitle}{\emph{Conference on learning theory}}. PMLR, \bibinfo{pages}{317--337}.
\newblock


\bibitem[\protect\citeauthoryear{Douze, Guzhva, Deng, Johnson, Szilvasy, Mazar{\'e}, Lomeli, Hosseini, and J{\'e}gou}{Douze et~al\mbox{.}}{2025}]%
        {douze2025faiss}
\bibfield{author}{\bibinfo{person}{Matthijs Douze}, \bibinfo{person}{Alexandr Guzhva}, \bibinfo{person}{Chengqi Deng}, \bibinfo{person}{Jeff Johnson}, \bibinfo{person}{Gergely Szilvasy}, \bibinfo{person}{Pierre-Emmanuel Mazar{\'e}}, \bibinfo{person}{Maria Lomeli}, \bibinfo{person}{Lucas Hosseini}, {and} \bibinfo{person}{Herv{\'e} J{\'e}gou}.} \bibinfo{year}{2025}\natexlab{}.
\newblock \showarticletitle{The faiss library}.
\newblock \bibinfo{journal}{\emph{IEEE Transactions on Big Data}} (\bibinfo{year}{2025}).
\newblock


\bibitem[\protect\citeauthoryear{Gao, Gou, Xu, Yang, Long, and Wong}{Gao et~al\mbox{.}}{2025}]%
        {gao2025erabitq}
\bibfield{author}{\bibinfo{person}{Jianyang Gao}, \bibinfo{person}{Yutong Gou}, \bibinfo{person}{Yuexuan Xu}, \bibinfo{person}{Yongyi Yang}, \bibinfo{person}{Cheng Long}, {and} \bibinfo{person}{Raymond Chi-Wing Wong}.} \bibinfo{year}{2025}\natexlab{}.
\newblock \showarticletitle{Practical and asymptotically optimal quantization of high-dimensional vectors in euclidean space for approximate nearest neighbor search}.
\newblock \bibinfo{journal}{\emph{Proceedings of the ACM on Management of Data}} \bibinfo{volume}{3}, \bibinfo{number}{3} (\bibinfo{year}{2025}), \bibinfo{pages}{1--26}.
\newblock


\bibitem[\protect\citeauthoryear{Gao and Long}{Gao and Long}{2024}]%
        {gao2024rabitq}
\bibfield{author}{\bibinfo{person}{Jianyang Gao} {and} \bibinfo{person}{Cheng Long}.} \bibinfo{year}{2024}\natexlab{}.
\newblock \showarticletitle{RaBitQ: Quantizing high-dimensional vectors with a theoretical error bound for approximate nearest neighbor search}.
\newblock \bibinfo{journal}{\emph{Proceedings of the ACM on Management of Data}} \bibinfo{volume}{2}, \bibinfo{number}{3} (\bibinfo{year}{2024}), \bibinfo{pages}{1--27}.
\newblock


\bibitem[\protect\citeauthoryear{Gao, Xiong, Gao, Jia, Pan, Bi, Dai, Sun, Wang, and Wang}{Gao et~al\mbox{.}}{2024}]%
        {gao2024ragsurvey}
\bibfield{author}{\bibinfo{person}{Yunfan Gao}, \bibinfo{person}{Yun Xiong}, \bibinfo{person}{Xinyu Gao}, \bibinfo{person}{Kangxiang Jia}, \bibinfo{person}{Jinliu Pan}, \bibinfo{person}{Yuxi Bi}, \bibinfo{person}{Yi Dai}, \bibinfo{person}{Jiawei Sun}, \bibinfo{person}{Meng Wang}, {and} \bibinfo{person}{Haofen Wang}.} \bibinfo{year}{2024}\natexlab{}.
\newblock \bibinfo{title}{Retrieval-Augmented Generation for Large Language Models: A Survey}.
\newblock
\newblock
\showeprint[arxiv]{2312.10997}~[cs.CL]
\urldef\tempurl%
\url{https://arxiv.org/abs/2312.10997}
\showURL{%
\tempurl}


\bibitem[\protect\citeauthoryear{Ge, He, Ke, and Sun}{Ge et~al\mbox{.}}{2014}]%
        {ge2014opq}
\bibfield{author}{\bibinfo{person}{Tiezheng Ge}, \bibinfo{person}{Kaiming He}, \bibinfo{person}{Qifa Ke}, {and} \bibinfo{person}{Jian Sun}.} \bibinfo{year}{2014}\natexlab{}.
\newblock \showarticletitle{Optimized Product Quantization}.
\newblock \bibinfo{journal}{\emph{IEEE Transactions on Pattern Analysis and Machine Intelligence}} \bibinfo{volume}{36}, \bibinfo{number}{4} (\bibinfo{year}{2014}), \bibinfo{pages}{744--755}.
\newblock
\urldef\tempurl%
\url{https://doi.org/10.1109/TPAMI.2013.240}
\showDOI{\tempurl}


\bibitem[\protect\citeauthoryear{Gray and Neuhoff}{Gray and Neuhoff}{1998}]%
        {gray1998quantization}
\bibfield{author}{\bibinfo{person}{Robert~M. Gray} {and} \bibinfo{person}{David~L. Neuhoff}.} \bibinfo{year}{1998}\natexlab{}.
\newblock \showarticletitle{Quantization}.
\newblock \bibinfo{journal}{\emph{IEEE Transactions on Information Theory}} \bibinfo{volume}{44}, \bibinfo{number}{6} (\bibinfo{year}{1998}), \bibinfo{pages}{2325--2383}.
\newblock


\bibitem[\protect\citeauthoryear{Guo, Sun, Lindgren, Geng, Simcha, Chern, and Kumar}{Guo et~al\mbox{.}}{2020}]%
        {guo2020scann}
\bibfield{author}{\bibinfo{person}{Ruiqi Guo}, \bibinfo{person}{Philip Sun}, \bibinfo{person}{Erik Lindgren}, \bibinfo{person}{Quan Geng}, \bibinfo{person}{David Simcha}, \bibinfo{person}{Felix Chern}, {and} \bibinfo{person}{Sanjiv Kumar}.} \bibinfo{year}{2020}\natexlab{}.
\newblock \showarticletitle{Accelerating large-scale inference with anisotropic vector quantization}. In \bibinfo{booktitle}{\emph{International Conference on Machine Learning}}. PMLR, \bibinfo{pages}{3887--3896}.
\newblock


\bibitem[\protect\citeauthoryear{He, Liu, Liu, Feng, Lu, Gan, Peng, Du, Yang, Liu, et~al\mbox{.}}{He et~al\mbox{.}}{2026}]%
        {he2026liftquant}
\bibfield{author}{\bibinfo{person}{Liulu He}, \bibinfo{person}{XuanAng Liu}, \bibinfo{person}{Juntao Liu}, \bibinfo{person}{Taolue Feng}, \bibinfo{person}{Ting Lu}, \bibinfo{person}{Chunsheng Gan}, \bibinfo{person}{Zhiyv Peng}, \bibinfo{person}{Yuan Du}, \bibinfo{person}{Huanrui Yang}, \bibinfo{person}{Yijiang Liu}, {et~al\mbox{.}}} \bibinfo{year}{2026}\natexlab{}.
\newblock \showarticletitle{LiftQuant: Continuous Bit-Width LLM via Dimensional Lifting and Projection}.
\newblock \bibinfo{journal}{\emph{arXiv preprint arXiv:2606.04050}} (\bibinfo{year}{2026}).
\newblock


\bibitem[\protect\citeauthoryear{Jayaram~Subramanya, Devvrit, Simhadri, Krishnawamy, and Kadekodi}{Jayaram~Subramanya et~al\mbox{.}}{2019}]%
        {jayaram2019diskann}
\bibfield{author}{\bibinfo{person}{Suhas Jayaram~Subramanya}, \bibinfo{person}{Fnu Devvrit}, \bibinfo{person}{Harsha~Vardhan Simhadri}, \bibinfo{person}{Ravishankar Krishnawamy}, {and} \bibinfo{person}{Rohan Kadekodi}.} \bibinfo{year}{2019}\natexlab{}.
\newblock \showarticletitle{DiskANN: Fast accurate billion-point nearest neighbor search on a single node}.
\newblock \bibinfo{journal}{\emph{Advances in neural information processing Systems}}  \bibinfo{volume}{32} (\bibinfo{year}{2019}).
\newblock


\bibitem[\protect\citeauthoryear{J{e}gou, Tavenard, Douze, and Amsaleg}{J{e}gou et~al\mbox{.}}{2011}]%
        {jegou2011bigann}
\bibfield{author}{\bibinfo{person}{Herv{\'{e}} J{e}gou}, \bibinfo{person}{Romain Tavenard}, \bibinfo{person}{Matthijs Douze}, {and} \bibinfo{person}{Laurent Amsaleg}.} \bibinfo{year}{2011}\natexlab{}.
\newblock \showarticletitle{Searching in One Billion Vectors: Re-evaluating the State of the Art}. In \bibinfo{booktitle}{\emph{Proceedings of the IEEE Conference on Computer Vision and Pattern Recognition ({CVPR})}}. \bibinfo{pages}{1817--1824}.
\newblock


\bibitem[\protect\citeauthoryear{Jégou, Douze, and Schmid}{Jégou et~al\mbox{.}}{2011}]%
        {jegou2011pq}
\bibfield{author}{\bibinfo{person}{Herve Jégou}, \bibinfo{person}{Matthijs Douze}, {and} \bibinfo{person}{Cordelia Schmid}.} \bibinfo{year}{2011}\natexlab{}.
\newblock \showarticletitle{Product Quantization for Nearest Neighbor Search}.
\newblock \bibinfo{journal}{\emph{IEEE Transactions on Pattern Analysis and Machine Intelligence}} \bibinfo{volume}{33}, \bibinfo{number}{1} (\bibinfo{year}{2011}), \bibinfo{pages}{117--128}.
\newblock
\urldef\tempurl%
\url{https://doi.org/10.1109/TPAMI.2010.57}
\showDOI{\tempurl}


\bibitem[\protect\citeauthoryear{Karpukhin, Oğuz, Min, Lewis, Wu, Edunov, Chen, and tau Yih}{Karpukhin et~al\mbox{.}}{2020}]%
        {karpukhin2020densepassage}
\bibfield{author}{\bibinfo{person}{Vladimir Karpukhin}, \bibinfo{person}{Barlas Oğuz}, \bibinfo{person}{Sewon Min}, \bibinfo{person}{Patrick Lewis}, \bibinfo{person}{Ledell Wu}, \bibinfo{person}{Sergey Edunov}, \bibinfo{person}{Danqi Chen}, {and} \bibinfo{person}{Wen tau Yih}.} \bibinfo{year}{2020}\natexlab{}.
\newblock \bibinfo{title}{Dense Passage Retrieval for Open-Domain Question Answering}.
\newblock
\newblock
\showeprint[arxiv]{2004.04906}~[cs.CL]
\urldef\tempurl%
\url{https://arxiv.org/abs/2004.04906}
\showURL{%
\tempurl}


\bibitem[\protect\citeauthoryear{Kuffo, Tsakalidou, De~Viti, Angel, I\v{s}a, and Lenhardt}{Kuffo et~al\mbox{.}}{2026}]%
        {kuffo2026semanticrecallvectorsearch}
\bibfield{author}{\bibinfo{person}{Leonardo Kuffo}, \bibinfo{person}{Ioanna Tsakalidou}, \bibinfo{person}{Roberta De~Viti}, \bibinfo{person}{Albert Angel}, \bibinfo{person}{Ji\v{r}\'{i} I\v{s}a}, {and} \bibinfo{person}{Rastislav Lenhardt}.} \bibinfo{year}{2026}\natexlab{}.
\newblock \showarticletitle{Semantic Recall for Vector Search}. In \bibinfo{booktitle}{\emph{Proceedings of the 49th International ACM SIGIR Conference on Research and Development in Information Retrieval}} (Melbourne, VIC, Australia) \emph{(\bibinfo{series}{SIGIR '26})}. \bibinfo{publisher}{Association for Computing Machinery}, \bibinfo{address}{New York, NY, USA}.
\newblock
\urldef\tempurl%
\url{https://doi.org/10.1145/3805712.3809894}
\showDOI{\tempurl}


\bibitem[\protect\citeauthoryear{Lee, Chavan, Nguyen, Huang, Jiang, Panda, Mertens, and Shukla}{Lee et~al\mbox{.}}{2026}]%
        {lee2026orbitquantdataagnosticquantizationimage}
\bibfield{author}{\bibinfo{person}{Donghyun Lee}, \bibinfo{person}{Jitesh Chavan}, \bibinfo{person}{Duy Nguyen}, \bibinfo{person}{Sam Huang}, \bibinfo{person}{Liming Jiang}, \bibinfo{person}{Priyadarshini Panda}, \bibinfo{person}{Timo Mertens}, {and} \bibinfo{person}{Saurabh Shukla}.} \bibinfo{year}{2026}\natexlab{}.
\newblock \bibinfo{title}{OrbitQuant: Data-Agnostic Quantization for Image and Video Diffusion Transformers}.
\newblock
\newblock
\showeprint[arxiv]{2607.02461}~[cs.CV]
\urldef\tempurl%
\url{https://arxiv.org/abs/2607.02461}
\showURL{%
\tempurl}


\bibitem[\protect\citeauthoryear{Li, Liu, Satapathy, Wang, and Cambria}{Li et~al\mbox{.}}{2023}]%
        {li2023recommendersystems}
\bibfield{author}{\bibinfo{person}{Yang Li}, \bibinfo{person}{Kangbo Liu}, \bibinfo{person}{Ranjan Satapathy}, \bibinfo{person}{Suhang Wang}, {and} \bibinfo{person}{Erik Cambria}.} \bibinfo{year}{2023}\natexlab{}.
\newblock \bibinfo{title}{Recent Developments in Recommender Systems: A Survey}.
\newblock
\newblock
\showeprint[arxiv]{2306.12680}~[cs.IR]
\urldef\tempurl%
\url{https://arxiv.org/abs/2306.12680}
\showURL{%
\tempurl}


\bibitem[\protect\citeauthoryear{Liu, Yuan, Jin, Zhong, Xu, Braverman, Chen, and Hu}{Liu et~al\mbox{.}}{2024}]%
        {liu2024kivi}
\bibfield{author}{\bibinfo{person}{Zirui Liu}, \bibinfo{person}{Jiayi Yuan}, \bibinfo{person}{Hongye Jin}, \bibinfo{person}{Shaochen Zhong}, \bibinfo{person}{Zhaozhuo Xu}, \bibinfo{person}{Vladimir Braverman}, \bibinfo{person}{Beidi Chen}, {and} \bibinfo{person}{Xia Hu}.} \bibinfo{year}{2024}\natexlab{}.
\newblock \showarticletitle{Kivi: A tuning-free asymmetric 2bit quantization for kv cache}.
\newblock \bibinfo{journal}{\emph{arXiv preprint arXiv:2402.02750}} (\bibinfo{year}{2024}).
\newblock


\bibitem[\protect\citeauthoryear{Ma, Zhang, Han, Yu, Wang, Ning, Zhang, Xu, Li, Qiao, Ju, Chen, Wang, Liu, Wang, Wang, Fu, Liu, Zhou, and Lu}{Ma et~al\mbox{.}}{2026}]%
        {ma2026vectordb}
\bibfield{author}{\bibinfo{person}{Le Ma}, \bibinfo{person}{Ran Zhang}, \bibinfo{person}{Yikun Han}, \bibinfo{person}{Shirui Yu}, \bibinfo{person}{Zaitian Wang}, \bibinfo{person}{Zhiyuan Ning}, \bibinfo{person}{Jinghan Zhang}, \bibinfo{person}{Ping Xu}, \bibinfo{person}{Pengjiang Li}, \bibinfo{person}{Ziyue Qiao}, \bibinfo{person}{Wei Ju}, \bibinfo{person}{Chong Chen}, \bibinfo{person}{Dongjie Wang}, \bibinfo{person}{Kunpeng Liu}, \bibinfo{person}{Pengyang Wang}, \bibinfo{person}{Pengfei Wang}, \bibinfo{person}{Yanjie Fu}, \bibinfo{person}{Chunjiang Liu}, \bibinfo{person}{Yuanchun Zhou}, {and} \bibinfo{person}{Chang-Tien Lu}.} \bibinfo{year}{2026}\natexlab{}.
\newblock \bibinfo{title}{A Comprehensive Survey on Vector Database: Storage and Retrieval Technique, Challenge}.
\newblock
\newblock
\showeprint[arxiv]{2310.11703}~[cs.DB]
\urldef\tempurl%
\url{https://arxiv.org/abs/2310.11703}
\showURL{%
\tempurl}


\bibitem[\protect\citeauthoryear{Monir, Lau, Yang, and Zhao}{Monir et~al\mbox{.}}{2024}]%
        {monir2024vectorsearch}
\bibfield{author}{\bibinfo{person}{Solmaz~Seyed Monir}, \bibinfo{person}{Irene Lau}, \bibinfo{person}{Shubing Yang}, {and} \bibinfo{person}{Dongfang Zhao}.} \bibinfo{year}{2024}\natexlab{}.
\newblock \bibinfo{title}{VectorSearch: Enhancing Document Retrieval with Semantic Embeddings and Optimized Search}.
\newblock
\newblock
\showeprint[arxiv]{2409.17383}~[cs.IR]
\urldef\tempurl%
\url{https://arxiv.org/abs/2409.17383}
\showURL{%
\tempurl}


\bibitem[\protect\citeauthoryear{Novikov, Vũ, Eisenberger, Dupont, Huang, Wagner, Shirobokov, Kozlovskii, Ruiz, Mehrabian, Kumar, See, Chaudhuri, Holland, Davies, Nowozin, Kohli, and Balog}{Novikov et~al\mbox{.}}{2025}]%
        {Novikov2025AlphaEvolve}
\bibfield{author}{\bibinfo{person}{Alexander Novikov}, \bibinfo{person}{Ngân Vũ}, \bibinfo{person}{Marvin Eisenberger}, \bibinfo{person}{Emilien Dupont}, \bibinfo{person}{Po-Sen Huang}, \bibinfo{person}{Adam~Zsolt Wagner}, \bibinfo{person}{Sergey Shirobokov}, \bibinfo{person}{Borislav Kozlovskii}, \bibinfo{person}{Francisco J.~R. Ruiz}, \bibinfo{person}{Abbas Mehrabian}, \bibinfo{person}{M.~Pawan Kumar}, \bibinfo{person}{Abigail See}, \bibinfo{person}{Swarat Chaudhuri}, \bibinfo{person}{George Holland}, \bibinfo{person}{Alex Davies}, \bibinfo{person}{Sebastian Nowozin}, \bibinfo{person}{Pushmeet Kohli}, {and} \bibinfo{person}{Matej Balog}.} \bibinfo{year}{2025}\natexlab{}.
\newblock \showarticletitle{AlphaEvolve: A coding agent for scientific and algorithmic discovery}.
\newblock \bibinfo{journal}{\emph{arXiv}} (\bibinfo{year}{2025}).
\newblock
\urldef\tempurl%
\url{https://doi.org/10.48550/arxiv.2506.13131}
\showDOI{\tempurl}


\bibitem[\protect\citeauthoryear{Pennington, Socher, and Manning}{Pennington et~al\mbox{.}}{2014}]%
        {pennington2014glove}
\bibfield{author}{\bibinfo{person}{Jeffrey Pennington}, \bibinfo{person}{Richard Socher}, {and} \bibinfo{person}{Christopher Manning}.} \bibinfo{year}{2014}\natexlab{}.
\newblock \showarticletitle{{G}lo{V}e: Global Vectors for Word Representation}. In \bibinfo{booktitle}{\emph{Proceedings of the 2014 Conference on Empirical Methods in Natural Language Processing ({EMNLP})}}, \bibfield{editor}{\bibinfo{person}{Alessandro Moschitti}, \bibinfo{person}{Bo~Pang}, {and} \bibinfo{person}{Walter Daelemans}} (Eds.). \bibinfo{publisher}{Association for Computational Linguistics}, \bibinfo{address}{Doha, Qatar}, \bibinfo{pages}{1532--1543}.
\newblock
\urldef\tempurl%
\url{https://doi.org/10.3115/v1/D14-1162}
\showDOI{\tempurl}


\bibitem[\protect\citeauthoryear{Raza, Rahman, Kamawal, Toroghi, Raval, Navah, and Kazemeini}{Raza et~al\mbox{.}}{2025}]%
        {raza2025recommendersystems}
\bibfield{author}{\bibinfo{person}{Shaina Raza}, \bibinfo{person}{Mizanur Rahman}, \bibinfo{person}{Safiullah Kamawal}, \bibinfo{person}{Armin Toroghi}, \bibinfo{person}{Ananya Raval}, \bibinfo{person}{Farshad Navah}, {and} \bibinfo{person}{Amirmohammad Kazemeini}.} \bibinfo{year}{2025}\natexlab{}.
\newblock \bibinfo{title}{A Comprehensive Review of Recommender Systems: Transitioning from Theory to Practice}.
\newblock
\newblock
\showeprint[arxiv]{2407.13699}~[cs.IR]
\urldef\tempurl%
\url{https://arxiv.org/abs/2407.13699}
\showURL{%
\tempurl}


\bibitem[\protect\citeauthoryear{Sharma}{Sharma}{2025}]%
        {sharma2025ragsurvey}
\bibfield{author}{\bibinfo{person}{Chaitanya Sharma}.} \bibinfo{year}{2025}\natexlab{}.
\newblock \bibinfo{title}{Retrieval-Augmented Generation: A Comprehensive Survey of Architectures, Enhancements, and Robustness Frontiers}.
\newblock
\newblock
\showeprint[arxiv]{2506.00054}~[cs.IR]
\urldef\tempurl%
\url{https://arxiv.org/abs/2506.00054}
\showURL{%
\tempurl}


\bibitem[\protect\citeauthoryear{Shi, Gao, Xia, Feh{\'e}r, and Long}{Shi et~al\mbox{.}}{2026}]%
        {shi2026ivfrabitq}
\bibfield{author}{\bibinfo{person}{Jifan Shi}, \bibinfo{person}{Jianyang Gao}, \bibinfo{person}{James Xia}, \bibinfo{person}{Tam{\'a}s~B{\'e}la Feh{\'e}r}, {and} \bibinfo{person}{Cheng Long}.} \bibinfo{year}{2026}\natexlab{}.
\newblock \showarticletitle{GPU-Native Approximate Nearest Neighbor Search with IVF-RaBitQ: Fast Index Build and Search}.
\newblock \bibinfo{journal}{\emph{arXiv preprint arXiv:2602.23999}} (\bibinfo{year}{2026}).
\newblock


\bibitem[\protect\citeauthoryear{Sun, Simcha, Dopson, Guo, and Kumar}{Sun et~al\mbox{.}}{2023}]%
        {sun2023soar}
\bibfield{author}{\bibinfo{person}{Philip Sun}, \bibinfo{person}{David Simcha}, \bibinfo{person}{Dave Dopson}, \bibinfo{person}{Ruiqi Guo}, {and} \bibinfo{person}{Sanjiv Kumar}.} \bibinfo{year}{2023}\natexlab{}.
\newblock \showarticletitle{SOAR: Improved Indexing for Approximate Nearest Neighbor Search}. In \bibinfo{booktitle}{\emph{Advances in Neural Information Processing Systems}}, Vol.~\bibinfo{volume}{36}. \bibinfo{publisher}{Curran Associates, Inc.}, \bibinfo{pages}{3189--3204}.
\newblock


\bibitem[\protect\citeauthoryear{Vallaeys, Muckley, Verbeek, and Douze}{Vallaeys et~al\mbox{.}}{2025}]%
        {vallaeys2025qinco2}
\bibfield{author}{\bibinfo{person}{Th{\'e}ophane Vallaeys}, \bibinfo{person}{Matthew~J Muckley}, \bibinfo{person}{Jakob Verbeek}, {and} \bibinfo{person}{Matthijs Douze}.} \bibinfo{year}{2025}\natexlab{}.
\newblock \showarticletitle{Qinco2: Vector compression and search with improved implicit neural codebooks}. In \bibinfo{booktitle}{\emph{International Conference on Learning Representations}}, Vol.~\bibinfo{volume}{2025}. \bibinfo{pages}{85467--85484}.
\newblock


\bibitem[\protect\citeauthoryear{Vargaftik, Ben-Basat, Portnoy, Mendelson, and Ben-Itzhak}{Vargaftik et~al\mbox{.}}{2022}]%
        {vargaftik2022eden}
\bibfield{author}{\bibinfo{person}{Shay Vargaftik}, \bibinfo{person}{Ran Ben-Basat}, \bibinfo{person}{Amit Portnoy}, \bibinfo{person}{Gal Mendelson}, {and} \bibinfo{person}{Yaniv Ben-Itzhak}.} \bibinfo{year}{2022}\natexlab{}.
\newblock \showarticletitle{EDEN: Communication-Efficient and Robust Distributed Mean Estimation for Federated Learning}. In \bibinfo{booktitle}{\emph{International Conference on Machine Learning (ICML)}} \emph{(\bibinfo{series}{Proceedings of Machine Learning Research})}, Vol.~\bibinfo{volume}{162}. \bibinfo{pages}{21984--22014}.
\newblock
\urldef\tempurl%
\url{https://proceedings.mlr.press/v162/vargaftik22a/vargaftik22a.pdf}
\showURL{%
\tempurl}


\bibitem[\protect\citeauthoryear{Vecchiato}{Vecchiato}{2024}]%
        {vecchiato2024learntrepresentatives}
\bibfield{author}{\bibinfo{person}{Thomas Vecchiato}.} \bibinfo{year}{2024}\natexlab{}.
\newblock \bibinfo{title}{Learning Cluster Representatives for Approximate Nearest Neighbor Search}.
\newblock
\newblock
\showeprint[arxiv]{2412.05921}~[cs.IR]
\urldef\tempurl%
\url{https://arxiv.org/abs/2412.05921}
\showURL{%
\tempurl}


\bibitem[\protect\citeauthoryear{Wang, Yi, Mao, et~al\mbox{.}}{Wang et~al\mbox{.}}{2021}]%
        {wang2021milvus}
\bibfield{author}{\bibinfo{person}{Jianguo Wang}, \bibinfo{person}{Xiaomeng Yi}, \bibinfo{person}{Rui Mao}, {et~al\mbox{.}}} \bibinfo{year}{2021}\natexlab{}.
\newblock \showarticletitle{Milvus: A Purpose-Built Vector Data Management System}. In \bibinfo{booktitle}{\emph{Proceedings of the 2021 International Conference on Management of Data (SIGMOD '21)}}. \bibinfo{publisher}{Association for Computing Machinery}, \bibinfo{pages}{2614--2627}.
\newblock
\urldef\tempurl%
\url{https://doi.org/10.1145/3448016.3457550}
\showDOI{\tempurl}


\bibitem[\protect\citeauthoryear{Wang, Yu, Zhang, and Zhang}{Wang et~al\mbox{.}}{2026}]%
        {wang2026pyramidpd}
\bibfield{author}{\bibinfo{person}{Yang Wang}, \bibinfo{person}{Lu Yu}, \bibinfo{person}{Jinbin Zhang}, {and} \bibinfo{person}{Qiyuan Zhang}.} \bibinfo{year}{2026}\natexlab{}.
\newblock \showarticletitle{Pyramid Product Quantization for Approximate Nearest Neighbor Search}.
\newblock \bibinfo{journal}{\emph{Applied Sciences}} \bibinfo{volume}{16}, \bibinfo{number}{2} (\bibinfo{year}{2026}), \bibinfo{pages}{853}.
\newblock


\bibitem[\protect\citeauthoryear{Widrow}{Widrow}{1956}]%
        {widrow1956study}
\bibfield{author}{\bibinfo{person}{Bernard Widrow}.} \bibinfo{year}{1956}\natexlab{}.
\newblock \showarticletitle{A study of rough amplitude quantization by means of Nyquist sampling theory}.
\newblock \bibinfo{journal}{\emph{IRE Transactions on Circuit Theory}} \bibinfo{volume}{3}, \bibinfo{number}{4} (\bibinfo{year}{1956}), \bibinfo{pages}{266--276}.
\newblock


\bibitem[\protect\citeauthoryear{Widrow and Koll{\'a}r}{Widrow and Koll{\'a}r}{2008}]%
        {widrow2008quantization}
\bibfield{author}{\bibinfo{person}{Bernard Widrow} {and} \bibinfo{person}{Istv{\'a}n Koll{\'a}r}.} \bibinfo{year}{2008}\natexlab{}.
\newblock \bibinfo{booktitle}{\emph{Quantization Noise: Roundoff Error in Digital Computation, Signal Processing, Control and Communications}}.
\newblock \bibinfo{publisher}{Cambridge University Press}, \bibinfo{address}{Cambridge}.
\newblock
\showISBNx{9780521886717}


\bibitem[\protect\citeauthoryear{Wu et~al\mbox{.}}{Wu et~al\mbox{.}}{2025}]%
        {wu2025polarquant}
\bibfield{author}{\bibinfo{person}{Songhao Wu} {et~al\mbox{.}}} \bibinfo{year}{2025}\natexlab{}.
\newblock \showarticletitle{PolarQuant: Leveraging Polar Transformation for Efficient Key Cache Quantization and Decoding Acceleration}.
\newblock \bibinfo{journal}{\emph{arXiv preprint arXiv:2502.00527}} (\bibinfo{year}{2025}).
\newblock


\bibitem[\protect\citeauthoryear{Xiao, Lin, Seznec, Wu, Demouth, and Han}{Xiao et~al\mbox{.}}{2023}]%
        {xiao2023smoothquant}
\bibfield{author}{\bibinfo{person}{Guangxuan Xiao}, \bibinfo{person}{Ji Lin}, \bibinfo{person}{Mickael Seznec}, \bibinfo{person}{Hao Wu}, \bibinfo{person}{Julien Demouth}, {and} \bibinfo{person}{Song Han}.} \bibinfo{year}{2023}\natexlab{}.
\newblock \showarticletitle{Smoothquant: Accurate and efficient post-training quantization for large language models}. In \bibinfo{booktitle}{\emph{International conference on machine learning}}. PMLR, \bibinfo{pages}{38087--38099}.
\newblock


\bibitem[\protect\citeauthoryear{Xu, Ji, Zhu, and Che}{Xu et~al\mbox{.}}{2025}]%
        {xu2025crvq}
\bibfield{author}{\bibinfo{person}{Yuzhuang Xu}, \bibinfo{person}{Shiyu Ji}, \bibinfo{person}{Qingfu Zhu}, {and} \bibinfo{person}{Wanxiang Che}.} \bibinfo{year}{2025}\natexlab{}.
\newblock \showarticletitle{CRVQ: Channel-relaxed vector quantization for extreme compression of LLMs}.
\newblock \bibinfo{journal}{\emph{Transactions of the Association for Computational Linguistics}}  \bibinfo{volume}{13} (\bibinfo{year}{2025}), \bibinfo{pages}{1488--1506}.
\newblock


\bibitem[\protect\citeauthoryear{Zagitov, Molodtsov, and Beznosikov}{Zagitov et~al\mbox{.}}{2026}]%
        {zagitov2026harp}
\bibfield{author}{\bibinfo{person}{Artur Zagitov}, \bibinfo{person}{Gleb Molodtsov}, {and} \bibinfo{person}{Aleksandr Beznosikov}.} \bibinfo{year}{2026}\natexlab{}.
\newblock \showarticletitle{HARP: Hadamard-Preconditioned Adaptive Rotation Processor for Extreme LLM Quantization}.
\newblock \bibinfo{journal}{\emph{arXiv preprint arXiv:2605.29843}} (\bibinfo{year}{2026}).
\newblock


\bibitem[\protect\citeauthoryear{Zandieh, Daliri, Hadian, and Mirrokni}{Zandieh et~al\mbox{.}}{2026}]%
        {zandieh2026turboquant}
\bibfield{author}{\bibinfo{person}{Amir Zandieh}, \bibinfo{person}{Majid Daliri}, \bibinfo{person}{Majid Hadian}, {and} \bibinfo{person}{Vahab Mirrokni}.} \bibinfo{year}{2026}\natexlab{}.
\newblock \showarticletitle{TurboQuant: Online Vector Quantization with Near-optimal Distortion Rate}. In \bibinfo{booktitle}{\emph{International Conference on Learning Representations (ICLR)}}.
\newblock


\bibitem[\protect\citeauthoryear{Zandieh, Daliri, and Han}{Zandieh et~al\mbox{.}}{2025}]%
        {zandieh2025qjl}
\bibfield{author}{\bibinfo{person}{Amir Zandieh}, \bibinfo{person}{Majid Daliri}, {and} \bibinfo{person}{Insu Han}.} \bibinfo{year}{2025}\natexlab{}.
\newblock \showarticletitle{Qjl: 1-bit quantized jl transform for kv cache quantization with zero overhead}. In \bibinfo{booktitle}{\emph{Proceedings of the AAAI Conference on Artificial Intelligence}}, Vol.~\bibinfo{volume}{39}. \bibinfo{pages}{25805--25813}.
\newblock


\bibitem[\protect\citeauthoryear{Zhao, Zheng, Yi, Luan, Xie, Zhou, and Jensen}{Zhao et~al\mbox{.}}{2023}]%
        {zhao2023fargo}
\bibfield{author}{\bibinfo{person}{Xi Zhao}, \bibinfo{person}{Bolong Zheng}, \bibinfo{person}{Xiaomeng Yi}, \bibinfo{person}{Xiaofan Luan}, \bibinfo{person}{Charles Xie}, \bibinfo{person}{Xiaofang Zhou}, {and} \bibinfo{person}{Christian~S Jensen}.} \bibinfo{year}{2023}\natexlab{}.
\newblock \showarticletitle{FARGO: Fast maximum inner product search via global multi-probing}.
\newblock \bibinfo{journal}{\emph{Proceedings of the VLDB Endowment}} \bibinfo{volume}{16}, \bibinfo{number}{5} (\bibinfo{year}{2023}), \bibinfo{pages}{1100--1112}.
\newblock


\end{thebibliography}
